\pdfoutput=1

\documentclass[11pt]{article}

\usepackage[final]{acl}

\usepackage{times}
\usepackage{latexsym}

\usepackage[T1]{fontenc}

\usepackage[utf8]{inputenc}

\usepackage{microtype}

\usepackage{inconsolata}

\usepackage{graphicx}

\usepackage{amsmath}
\usepackage{amsthm}
\usepackage{booktabs}
\usepackage{multicol}
\usepackage{adjustbox}
\usepackage{subcaption}
\usepackage{multirow}
\usepackage{algpseudocode}
\usepackage{algorithm}
\usepackage{amssymb}
\usepackage{makecell}
\usepackage[multiple]{footmisc}
\usepackage{resizegather}
\usepackage{microtype}
\usepackage{afterpage}
\usepackage{enumitem}
\usepackage{hyperref}
\usepackage{dsfont}

\title{Analyzing and Evaluating Correlation Measures in NLG Meta-Evaluation}

\author{Mingqi Gao$^{*}$, Xinyu Hu$^{*}$, Li Lin, Xiaojun Wan \\
         Wangxuan Institute of Computer Technology, Peking University\\
         \texttt{\{gaomingqi,huxinyu,wanxiaojun\}@pku.edu.cn}\\
         \texttt{efsotr\_l@stu.pku.edu.cn}}

\begin{document}
\maketitle
\def\thefootnote{*}\footnotetext{Equal contribution.}\def\thefootnote{\arabic{footnote}}

\begin{abstract}
The correlation between NLG automatic evaluation metrics and human evaluation is often regarded as a critical criterion for assessing the capability of an evaluation metric. However, different grouping methods and correlation coefficients result in various types of correlation measures used in meta-evaluation. In specific evaluation scenarios, prior work often directly follows conventional measure settings, but the characteristics and differences between these measures have not gotten sufficient attention. Therefore, this paper analyzes 12 common correlation measures using a large amount of real-world data from six widely-used NLG evaluation datasets and 32 evaluation metrics, revealing that different measures indeed impact the meta-evaluation results. Furthermore, we propose three perspectives that reflect the capability of meta-evaluation: \textbf{discriminative power}, \textbf{ranking consistency}, and \textbf{sensitivity to score granularity}. We find that the measure using global grouping and Pearson correlation coefficient exhibits the best performance in both discriminative power and ranking consistency. Besides, the measures using system-level grouping or Kendall correlation are the least sensitive to score granularity.
\end{abstract}

\section{Introduction}
\label{sec:intro}
Automatic evaluation metrics (e.g., BLEU \citep{DBLP:conf/acl/PapineniRWZ02} and BERTScore \citep{DBLP:conf/iclr/ZhangKWWA20}) are widely used in natural language generation (NLG) evaluation to assess the quality of content generated by a system or model for a specific task. Furthermore, the evaluation of the performance of these evaluation metrics is referred to as meta-evaluation, which typically uses the correlation between the metrics and human evaluation as a crucial criterion, as human evaluation is generally considered the gold standard. However, the implementation of the correlation measure is not uniform because it involves two elements that have different possible selections: the grouping method of the evaluation scores (e.g., system level \citep{DBLP:conf/emnlp/BhandariGALN20} and input level \citep{DBLP:journals/tacl/DeutschDR21}) and the correlation coefficient (e.g., Pearson's $r$ and Spearman's $\rho$).

Our relevant preliminary experiments show that different correlation measures can indeed lead to different meta-evaluation results. However, prior studies have rarely paid attention to the relationships and differences between different measures; instead, they often simply follow the conventional practices of related work or authoritative competitions on evaluation. Some studies even do not clearly describe the correlation measure they used, particularly in terms of grouping methods, let alone explain the reasons for selecting that measure. Moreover, the correlation measures used in some authoritative competitions are constantly changing. For instance, WMT22 \citep{DBLP:conf/wmt/FreitagRMLSAKFLM22} used segment-level correlations that include three different grouping methods, whereas WMT21 \citep{DBLP:conf/wmt/FreitagRMLSFLB21} and WMT23 \citep{DBLP:conf/wmt/FreitagMLARTKBD23} only used one. These issues of non-transparency and inconsistency indicate that the correlation measure and meta-evaluation in NLG require more in-depth understanding.

On the other hand, large language models (LLMs) have been increasingly used in automatic evaluation, including both prompting proprietary LLMs for NLG evaluation \citep{DBLP:conf/emnlp/LiuIXWXZ23,DBLP:conf/acl/ChiangL23,DBLP:conf/eamt/KocmiF23} and fine-tuned LLM evaluators \citep{DBLP:journals/corr/abs-2310-11593,DBLP:conf/emnlp/XuWPSFWL23,DBLP:journals/corr/abs-2310-00752,DBLP:journals/corr/abs-2310-05470}. Unlike traditional continuous evaluation metrics, LLM-based evaluators typically output discrete scores and can assess on different evaluation scales based on flexible requirements (e.g., 1-5, 0-100). This introduces more evaluation ties and varying degrees of the granularity or discretization of evaluation scores, which may affect the fairness of comparisons in certain correlation measures \citep{DBLP:conf/emnlp/DeutschFF23}, making the already confusing selection of correlation measures even more complex.

However, it is by no means easy to strictly determine whether a particular correlation measure is reasonable, as it depends on the specific scenario and research objectives. For example, \citet{DBLP:conf/emnlp/DeutschFF23} believe that when using multidimensional quality metric (MQM) \citep{freitag2021experts} for machine translation evaluation, fine-grained tied scores should be trusted and used for meta-evaluation, while traditional Kendall correlation coefficients cannot handle these situations and are therefore unsuitable. However, in coarser-grained evaluations (such as Likert scale ratings from 1 to 5), ties may not be as reliable, rendering the above conclusion invalid. Therefore, this paper focuses on the comparison and analysis of different correlation measures and their meta-evaluation capabilities, primarily around the following three and characteristic perspectives and research questions:

\begin{itemize}
    \item \textbf{RQ1} (\S\ref{sec:DP}): For \textbf{discriminative power}, which correlation measures can more effectively distinguish between pairs of automatic evaluation metrics?
    \item \textbf{RQ2} (\S\ref{sec:RC}): For \textbf{ranking consistency}, which correlation measures can provide more stable rankings for a set of evaluation metrics?
    \item \textbf{RQ3} (\S\ref{sec:granularity}): For \textbf{granularity sensitivity}, which correlation measures can better handle different evaluation score granularity?
\end{itemize}

To achieve more comprehensive and realistic analyses, we collect six commonly-used NLG evaluation datasets, including 30 different subsets, and calculate and annotate the results of 32 different automatic evaluation metrics (including evaluators using LLMs such as GPT-4). Based on such large amounts of real data, we design specific testing algorithms to analyze the above three questions and summarize the corresponding conclusions. Our contributions are summed up as follows:

\begin{enumerate}
    \item We point out the necessity for a more thorough understanding of correlation measures in NLG meta-evaluation and demonstrate that the different selections of measures do affect meta-evaluation results.
    \item We propose three characteristics reflecting meta-evaluation capabilities as well as corresponding methods to analyze different correlation measures and experiment on a large amount of real-world data from six evaluation datasets and 32 evaluation metrics.
    \item Our experimental results show that the measure using global grouping and Pearson correlation exhibits the best meta-evaluation capabilities. We hope our work can deepen the understanding of correlation measures, thereby clarifying their usages in future research\footnote{Our code and data will be available on \url{https://github.com/kite99520/NLGCorrEval}.}.
\end{enumerate}

\section{Background}
In NLG evaluation, we usually focus on the quality of the output generated by a system or model for a given task and input. For example, in automatic news summarization, the input is a news article, and the output is a summary. There are two ways to evaluate the output: human evaluation and automatic evaluation, both typically expressed as scores. Human evaluation scores are considered the gold standard, and their consistency with automatic evaluation metrics is used to assess and compare the performance of different metrics, which is caculated by the specific correlation measure. This process is referred to as meta-evaluation and can be formalized as follows:

Assume that there are $N$ systems, $\{ s_i \}_{i=1}^N$ and $M$ inputs, $\{ d_j \}_{j=1}^M$. Each system $s_i$ generates an output $h_{ij}$ for each input $d_j$ and the human evaluation score for each output $h_{ij}$ is denoted as $z_{ij}$. These form a meta-evaluation dataset $D=\{ \{ d_j \}_{j=1}^{M}, \{(h_{ij}, z_{ij})\}_{i=1,j=1}^{N,M} \}$. In most meta-evaluation datasets, $N \ll M$, and generally, the range of $N$ is a few to dozens, while the range of $M$ is tens to thousands.

On the other hand, an automatic evaluation metric $m$ typically requires the input, output, and other related optional content, such as the reference for evaluation. For each output $h_{ij}$, the score given by the automatic evaluation metric is denoted as $x_{ij}$. If there are $K$ metrics to be evaluated in meta-evaluation, they are denoted as $\{ m_k \}_{k=1}^K$, and their output scores are denoted as $\{ x_{ij}^k \}_{k=1}^K$ (also denoted as matrices $\{X_k\}_{k=1}^K$).

The correlation $C(X, Z)$ between $\{ x_{ij} \}_{i=1,j=1}^{N,M}$ (i.e., $X$) and $\{ z_{ij} \}_{i=1,j=1}^{N,M}$ (i.e., $Z$) is used to evaluate the quality of a certain evaluation metric $m$. And there are multiple ways to measure this correlation, which can be divided into four categories based on the grouping method, where $c$ denotes a specific correlation coefficient, commonly Pearson's $r$, Spearman's $\rho$, and Kendall's $\tau$:

\begin{itemize}
    \item \textbf{Global Level}: Flatten matrices of evaluation scores into vectors and calculate the correlation coefficient between two $N \times M$-dimensional vectors, $C_{global}(X, Z) = c( (x_{ij})_{i=1,j=1}^{N,M}, (z_{ij})_{i=1,j=1}^{N,M})$.
    \item \textbf{Input Level}: For each input, calculate the correlation coefficient between two $N$-dimensional vectors, and then average the $M$ correlation coefficients, $C_{input}(X, Z) = \frac{1}{M} \sum_{j=1}^M c((x_{ij})_{i=1}^{N}, (z_{ij})_{i=1}^{N})$.
    \item \textbf{Item Level}\footnote{WMT22 \citep{DBLP:conf/wmt/FreitagRMLSAKFLM22} used this grouping as one of segment-level grouping methods. And we rename it "Item level" to avoid confusion.}: For each system, calculate the correlation coefficient between two $M$-dimensional vectors, and then average the $N$ correlation coefficients, $C_{item}(X, Z) = \frac{1}{N} \sum_{i=1}^N c((x_{ij})_{j=1}^{M}, (z_{ij})_{j=1}^{M})$.
    \item \textbf{System Level}\footnote{From a completeness perspective, there is another grouping method similar to system level, which first averages the scores of each input across $N$ systems, and then calculates the correlation coefficient between the two $M$-dimensional vectors. However, it may reflect the difficulty of inputs and has no significance in evaluation.}: First average the scores of each system across $M$ inputs, and then calculate the correlation coefficient between the two $N$-dimensional vectors, $C_{system}(X, Z) = c((\frac{1}{M}\sum_{j=1}^{M}x_{ij})_{i=1}^N,(\frac{1}{M}\sum_{j=1}^{M}z_{ij})_{i=1}^N)$.
\end{itemize}

It can be seen that the correlation measure includes two parts: the grouping method and the correlation coefficient. Considering the common four grouping methods and three correlation coefficients\footnote{All Kendall's $\tau$ in this paper refers to $\tau_b$. See Appendix \ref{app:correlation_coefficient} for the full definitions of three correlation coefficients.} mentioned above, there are $4 \times 3 = 12$ different correlation measures in total. For two evaluation metrics $m_1$ and $m_2$ and a specific correlation measure $C$, it is generally considered that $m_1$ outperforms $m_2$ if $C(X_1, Z) > C(X_2, Z)$. Furthermore, the ranking of $K$ metrics $\{ m_k \}_{k=1}^K$ can also be determined by comparisons within ${C(X_k, Z)}_{k=1}^K$. 

\begin{figure}[H]
  \includegraphics[width=1\linewidth]{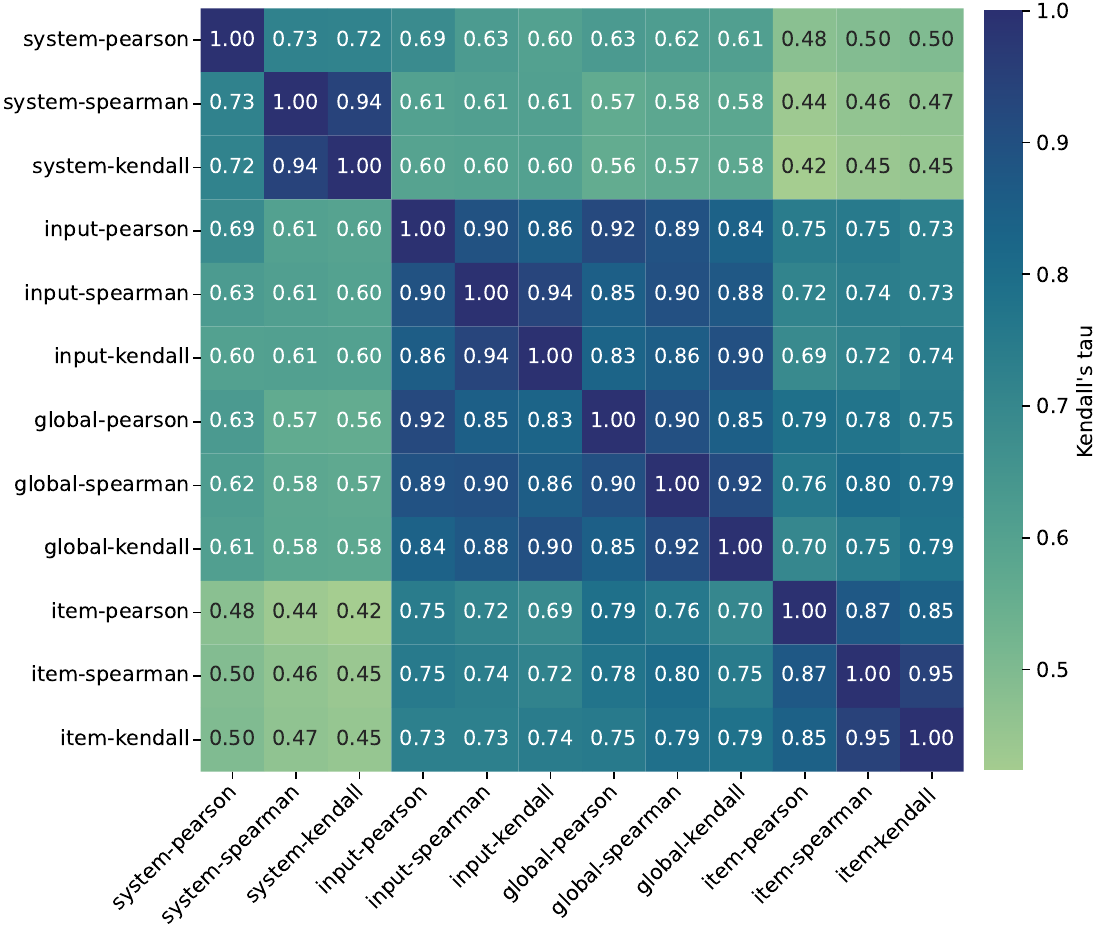}
  \caption{The consistency of evaluation metric rankings using different correlation measures on SummEval, calculated through Kendall's correlation coefficient.}
  \label{fig:preliminary}
\end{figure}

To verify whether different correlation measures would affect the results of meta-evaluation, we conduct preliminary experiments of 12 measures on common evaluation datasets. Specifically, for two correlation measures $C_1$ and $C_2$, as well as the evaluation metrics to be meta-evaluated $\{ m_k \}_{k=1}^K$, we calculate the consistency of metric rankings under the two measures using $\tau((C_1(X_k, Z))_{k=1}^K, (C_2(X_k, Z))_{k=1}^K)$. Since we are concerned with rankings here, we choose Kendall's $\tau$ as the correlation coefficient. The results on SummEval \citep{DBLP:journals/tacl/FabbriKMXSR21} are presented in Figure \ref{fig:preliminary}, with the details of data introduced in Section \ref{sec:data_preparation}. Any two different correlation measures lead to different meta-evaluations of metric rankings, with greater inconsistency when different grouping methods are used. Similar results are observed on other datasets, which are presented in Figures \ref{fig:hanna}-\ref{fig:wmt} in the appendix, indicating that the selection of correlation measures is indeed important and requires further analysis.

\begin{table*}[t]
\small
\centering
\resizebox{\linewidth}{!}{
\renewcommand{\arraystretch}{1.1}
\begin{tabular}{llcccc}
\toprule
\textbf{Task} & \textbf{Name} & \textbf{\#Subsets} & \textbf{\#Aspects} & \textbf{\#Systems} & \textbf{\#Inputs} \\ 
\midrule
Summarization & SummEval \citep{DBLP:journals/tacl/FabbriKMXSR21}  & 1 & 4 & 16 & 100 \\
Translation & WMT23-ZH2EN-NEWS \citep{DBLP:conf/wmt/FreitagMLARTKBD23} & 1 & 1 & 16 & 376 \\ 
Story Generation & HANNA \citep{DBLP:conf/coling/ChhunCSC22} & 1 & 6 & 6 & 60 \\ 
Story Generation & MANS\citep{DBLP:conf/acl/GuanZFLDMFH20} & 2 & 1 & 5 & 200 \\ 
Dialogue & USR \citep{DBLP:conf/acl/MehriE20}  & 2 & 6 & 5 & 60 \\ 
Data-to-text & WebNLG2020 \citep{webnlg-2020-international} & 1 & 5 & 16 & 178 \\ 
\bottomrule
\end{tabular}
}
\caption{Information and statistics of different evaluation datasets.}
\label{tab:dataset_meta_info}
\end{table*}

\section{Meta-Evaluation Data}
\label{sec:data_preparation}

For more comprehensive and realistic analyses on correlation measures, we collect six widely used evaluation datasets, as well as the results of 32 common automatic evaluation metrics on them. Our subsequent experiments on three research questions are based on this large-scale real evaluation data.

\subsection{Datasets}
As shown in Table \ref{tab:dataset_meta_info}, we select and preprocess six common evaluation datasets from five typical NLG tasks: summarization, story generation, dialogue, data-to-text, and translation. Due to the large volume of WMT23 data, we only use news domain data from ZH2EN. Following convention \citep{DBLP:conf/acl/GuanZFLDMFH20,DBLP:conf/acl/MehriE20}, we split the original datasets according to subsets and aspects, resulting in a total of 30 meta-evaluation datasets. Since our primary focus is on analyzing the general characteristics of correlation measures across different scenarios rather than on specific datasets, we label these datasets as D1-D30 for brevity and will summarize their overall performance in subsequent experiments. More details and the correspondence are shown in Table \ref{tab:subdata_parameter}.

\subsection{Automatic Evaluation Metrics}
\label{sec:metrics}
We select 14 common non-LLM evaluation metrics, including string-based BLEU \citep{DBLP:conf/acl/PapineniRWZ02}, ROUGE-(1,2,L)\footnote{ROUGE-(1,2,L) refers to three different variants of ROUGE. Similarly, BARTScore-(s-h,r-h,h-r) refers to three different ways of using BARTScore, typically regarded as different automatic evaluation metrics.} \citep{lin-2004-rouge}, CHRF \citep{DBLP:conf/wmt/Popovic15}, and model-based BERTScore-(p,r,f1) \citep{DBLP:conf/iclr/ZhangKWWA20}, MoverScore \citep{DBLP:conf/emnlp/ZhaoPLGME19}, BARTScore-(s-h, r-h, h-r) \citep{DBLP:conf/nips/YuanNL21}, BLEURT \citep{DBLP:conf/acl/SellamDP20}, and COMET \citep{DBLP:conf/emnlp/ReiSFL20}. For LLM-based evaluators, we employ 18 experimental settings to prompt proprietary LLMs to score outputs based on task descriptions and aspect definitions, resulting in 18 evaluation metrics: three different proprietary LLMs from OpenAI\footnote{\url{https://openai.com/api/}} (\texttt{gpt-3.5-turbo}, \texttt{gpt-4-turbo}, \texttt{gpt-4o}); different prompting strategies of three evaluation scales (1-5, 1-10, 0-100); and two sampling settings (temperature T=0 and sampling once, temperature T=1 and sampling ten times with results averaged). In total, there are $K=32$ automatic evaluation metrics, and more information including the detailed prompts and implementations are shown in Appendix \ref{app:evaluation_metrics}.

\section{Analyses of Correlation Measures}

To analyze and compare the meta-evaluation capabilities of different correlation measures, we consider three important perspectives: discriminative power, ranking consistency, and sensitivity to score granularity. In practice, correlation measures are mainly used to assess the performance of automatic evaluation metrics, with two core applications: comparing two metrics and ranking a set of metrics. For the former, the key lies in the discriminative power of the correlation measure, i.e., whether it can distinguish various metric pairs as effectively as possible. For the latter, the consistency of the correlation measure in ranking the same set of metrics is crucial, i.e., whether the ranking remains stable. In addition, recently emerging LLM-based evaluators feature a flexible and discrete scoring pattern similar to human evaluation, so the sensibility of correlation measures to varying levels of score granularity of the same metric should meet certain expectations. In the following subsections, we will design specific tests and conduct experimental analyses based on these three capability aspects to identify the characteristics of different correlation measures.

\subsection{Discriminative Power}
\label{sec:DP}

In the fields of information retrieval \citep{DBLP:conf/promisews/Sakai13} and recommendation systems \citep{DBLP:conf/recsys/AnelliNSPR19,DBLP:conf/sigir/AshkanM19,DBLP:journals/ir/ValcarceBPC20}, discriminative power is widely used to evaluate evaluation measures. Inspired by this, we adapted this method to evaluate correlation measures in NLG meta-evaluation.

\begin{figure*}[t]
  \includegraphics[width=\linewidth]{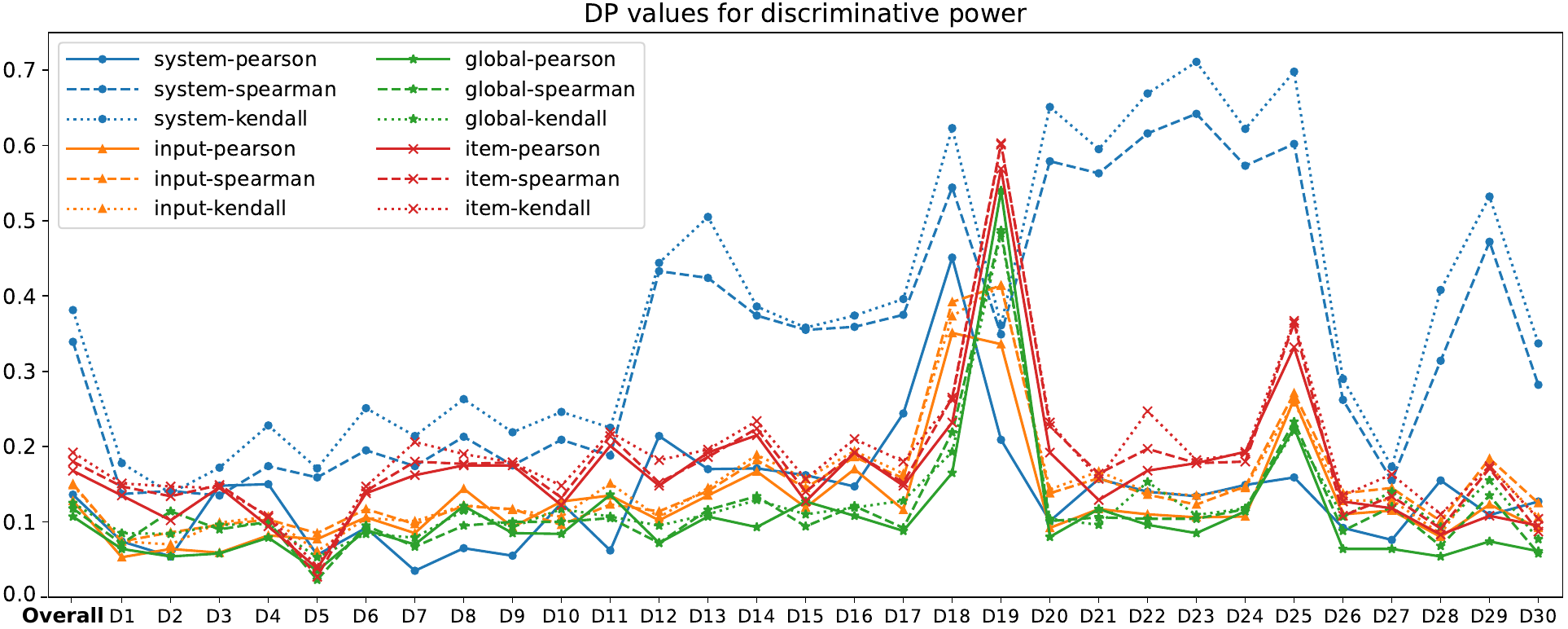}
  \caption {DP values of different correlation measures on all meta-evaluation datasets using the permutation test, \textbf{the lower the better}. Each column "Dn" shows the result on one dataset, which corresponds to the original dataset as shown in Table \ref{tab:subdata_parameter}. The first column presents the overall performance with the averaged results of all datasets.}
    \label{fig:DP}
\end{figure*}

Specifically, for a given correlation measure, a meta-evaluation dataset (including human scores $Z$), and the scores of $K$ automatic evaluation metrics on it $\{X_k\}_{k=1}^K$, we obtain the two-sided p-value for each pair of automatic evaluation metrics through hypothesis testing. The smaller the p-value, the more likely we are to reject the null hypothesis that there is no difference in the correlation values between the two metric scores with human scores. Thus, a highly discriminative correlation measure should yield many very small p-values. For convenience of comparison, similar to \citet{DBLP:journals/ir/ValcarceBPC20}, we define the DP value as the average of p-values of all metric pairs, ranging from 0 to 1, with smaller values indicating stronger discriminative power of a correlation measure. Algorithm \ref{alg:discrinimative_power} shows the pseudocode for calculating the DP value. Regarding the hypothesis testing methods used here, we refer to previous work and employ the Perm-Both algorithm proposed by \citet{DBLP:journals/tacl/DeutschDR21}\footnote{We used the code from \url{https://nlpstats.readthedocs.io/en/latest/}, with 1000 permutation samples.}. \citep{noreen1989computer}. It is a non-parametric test method that \citet{DBLP:journals/tacl/DeutschDR21} have shown to have a higher power in summarization meta-evaluation. Figure \ref{fig:DP} shows the DP values of correlation measures across all meta-evaluation datasets, and the overall performance and rankings are summarized in Table \ref{tab:DP_RC}. The complete results are shown in Table \ref{tab:DP} in the appendix.

In addition to giving an overall value, we can also look more closely at the discriminative power of different correlation measures by p-values curves, presented in Figures \ref{fig:p_value_curve_D1}-\ref{fig:p_value_curve_D30} in the appendix. After obtaining the p-values for each pair of evaluation metrics, we sort them in descending order. With the number of evaluation metric pairs on the x-axis and the p-values on the y-axis, we can plot the p-value curves of different correlation measures on a meta-evaluation dataset. The closer the curve is to the coordinate axis, the stronger the discriminative power of the corresponding correlation measure. The DP value numerically equals the area enclosed by the p-value curve and the coordinate axis divided by the number of metric pairs.

\begin{algorithm}[H]
{\small
\caption{Discriminative Power}
\label{alg:discrinimative_power}
 \textbf{Input:} $X_1,\dots,X_K,Z \in \mathbb{R}^{N\times M}$, $T \in \mathbb{N}$, $C$

\textbf{Output:} DP value
\begin{algorithmic}[0]
\State $v \gets 0$
\State $n \gets K\times (K-1) / 2$
\For{$i \in \{1,\dots,K-1\}$}
    \For{$j \in \{i+1,\dots,K\}$}
        \State $p_{ij} \gets$ \Call{PermutationTest}{$X_i, X_j, Z, T, C$} 
        \State $v\gets v + p_{ij}$
    \EndFor
\EndFor
\State \Return $v / n$
\end{algorithmic}

\begin{algorithmic}[0]
\Function{PermutationTest}{$X, Y, Z, T, C$}
\State $q\gets$ 0
\State $\delta \gets C(X, Z) - C(Y, Z)$
\For{$T$ iterations}
    \State $X_s, Y_s \gets$ empty $N \times M$ matrices
    \For{$(i, j) \in \{1, \dots, N\} \times \{1, \dots, M\}$}
        \If{random Boolean is true} 
            \State $X_s[i, j] \gets Y[i, j]$
            \State $Y_s[i, j] \gets X[i, j]$
        \Else 
            \State $X_s[i, j] \gets X[i, j]$
            \State $Y_s[i, j] \gets Y[i, j]$
        \EndIf
    \EndFor
    \State $\delta_s \gets C(X_s, Z) - C(Y_s, Z)$
    \If{ $|\delta_s| > |\delta|$}
        \State $q \gets q + 1$
    \EndIf
\EndFor
\State \Return $q / T$
\EndFunction
\end{algorithmic}
}
\end{algorithm}

\begin{figure*}[t]
  \includegraphics[width=\linewidth]{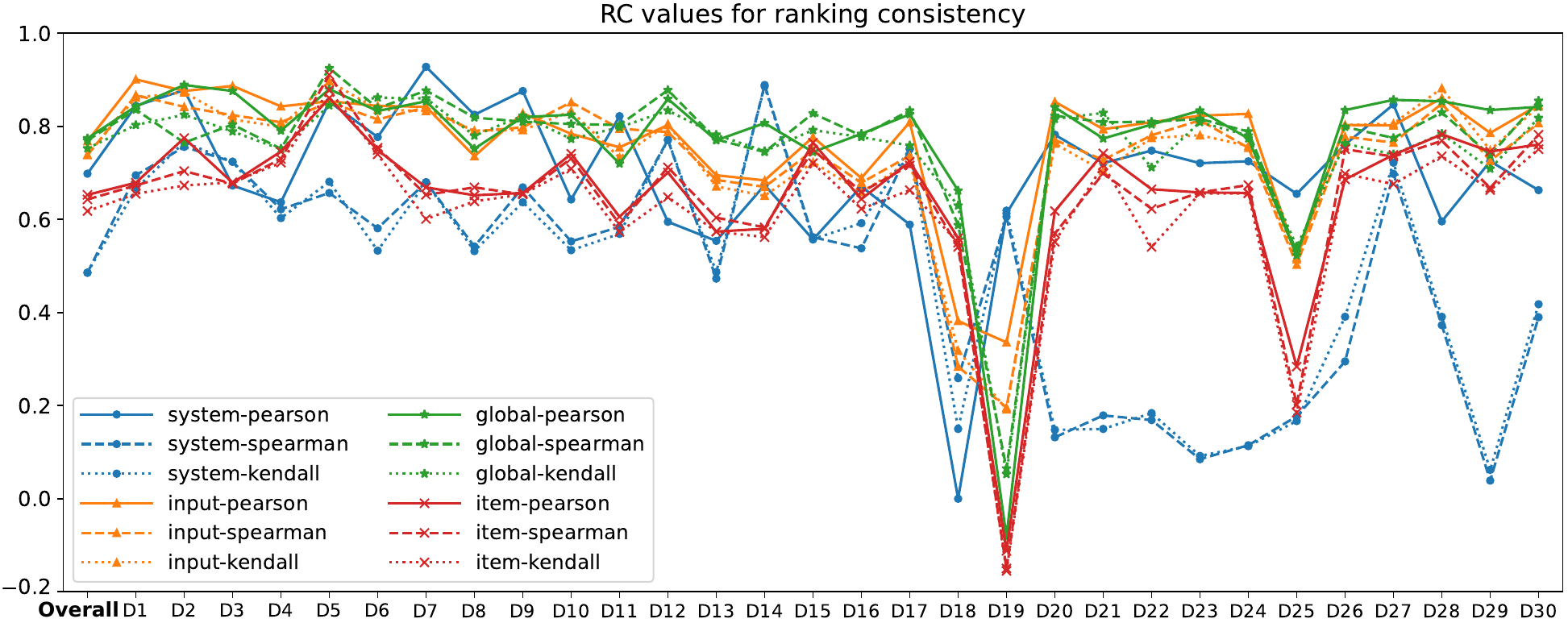}
  \caption {RC values of different correlation measures on all meta-evaluation datasets, \textbf{the higher the better}, with the representation of columns similar to Figure \ref{fig:DP}.}
    \label{fig:RC}
\end{figure*}

\paragraph{Takeaways} The overall discriminative power of different correlation measures can be summarized and ranked based on different grouping methods and correlation coefficients as follows:

\begin{itemize}
 \item Grouping: Global > Input > Item > System
 \item Correlation Coefficient: Pearson's $r$ > Spearman's $\rho$ > Kendall's $\tau$
\end{itemize}

\begin{algorithm}[H]
{\small
\caption{Ranking Consistency}
\label{alg:ranking_consistency}
 \textbf{Input:} $X_1,\dots,X_K,Z \in \mathbb{R}^{N\times M}$, 
 $T \in \mathbb{N}$ , $C$.

 \textbf{Output:} RC value
\begin{algorithmic}[0]
\State $v \gets 0$
\For{$T$ iterations}
    \State $M_1 \gets \lfloor M / 2 \rfloor$
    \State $M_2 \gets M - \lfloor M / 2 \rfloor$
    \State $D_1 \gets$ sample $\{1,\dots,M\}$ w/o repl. $M_1$ times
    \State $D_2 \gets \{1,\dots,M\} \setminus D_1$
    \State $R_1, R_2 \gets$ empty $K$-dimensional arrays
    \For{$k \in \{1,\dots,K\}$}
        \State $X^s_1, Z^s_1 \gets$ empty $N \times M_1$ matrices
        \State $X^s_2, Z^s_2 \gets$ empty $N \times M_2$ matrices
        \For{$i \in \{1,\dots,N\}$}
            \For{$j \in \{1,\dots,M_1\}$}
                \State $X^s_1[i, j] \gets X_k[i, D_1[j]]$
                \State $Z^s_1[i, j] \gets Z[i, D_1[j]]$               
            \EndFor
            \For{$j \in \{1,\dots,M_2\}$}
                \State $X^s_2[i, j] \gets X_k[i, D_2[j]]$
                \State $Z^s_2[i, j] \gets Z[i, D_2[j]]$    
            \EndFor
        \EndFor
        \State $R_1[k] \gets C(X^s_1, Z^s_1)$
        \State $R_2[k] \gets C(X^s_2, Z^s_2)$
    \EndFor
    \State $\tau^s \gets \tau(R_1, R_2)$
    \State $v \gets v + \tau^s$
\EndFor
\State \Return $v / T$
\end{algorithmic}
}
\end{algorithm}

\subsection{Ranking Consistency}
\label{sec:RC}

Inspired by the evaluation of different evaluation measures of ordinal classification \citep{DBLP:conf/acl/Sakai20} and information retrieval \citep{DBLP:conf/ecir/Sakai21}, 
for a given correlation measure, we randomly split the human scores and evaluation metric outputs in half, derive the rankings of the evaluation metrics on the two halves, and calculate the similarity of the two rankings using Kendall's $\tau$ as a measure of ranking consistency. We define the RC value as the mean results obtained from repeating this process $T=1000$ times. Algorithm \ref{alg:ranking_consistency} presents the pseudocode for the calculation. Figure \ref{fig:RC} and Table \ref{tab:DP_RC} shows the RC values of correlation measures across all meta-evaluation datasets, with the complete results presented in Table \ref{tab:RC} in the appendix. According to Table \ref{tab:DP_RC}, ranking consistency and discriminative power exhibit similar trends for correlation measures, meaning that correlation measures with high discriminative power generally also have high ranking consistency.

\begin{table}
\small
\resizebox{\linewidth}{!}{
\begin{tabular}{llll}
\toprule
Grouping & Correlation & DP value $\downarrow$ & RC value $\uparrow$ \\ 
\midrule
System & Pearson & 0.136 (5) & 0.698 (7) \\
System & Spearman & 0.339 (11) & 0.485 (12) \\
System & Kendall & 0.381 (12) & 0.486 (11) \\
Input & Pearson & 0.128 (4) & 0.768 (3) \\
Input & Spearman & 0.150 (7) & 0.738 (6) \\
Input & Kendall & 0.149 (6) & 0.739 (5) \\
Global & Pearson & \textbf{0.107 (1)} & \textbf{0.774 (1)} \\
Global & Spearman & 0.118 (2) & 0.770 (2) \\
Global & Kendall & 0.125 (3) & 0.753 (4) \\
Item & Pearson & 0.168 (8) & 0.653 (8) \\
Item & Spearman & 0.180 (9) & 0.643 (9) \\
Item & Kendall & 0.192 (10) & 0.618 (10) \\
\bottomrule
\end{tabular}
}
\caption{Overall performance and rankings (in brackets) of different correlation measures with the averaged results of all meta-evaluation datasets in terms of the discriminative power (DP) and ranking consistency (RC).} 
\label{tab:DP_RC}
\end{table}

\paragraph{Takeaways} The overall ranking consistency of different correlation measures can be summarized and ranked based on different grouping methods and correlation coefficients as follows:

\begin{itemize}
 \item Grouping: Global > Input > Item > System
 \item Correlation Coefficient: Pearson's $r$ > Spearman's $\rho$ > Kendall's $\tau$
\end{itemize}

\begin{figure*}[t]
  \includegraphics[width=0.49\linewidth]{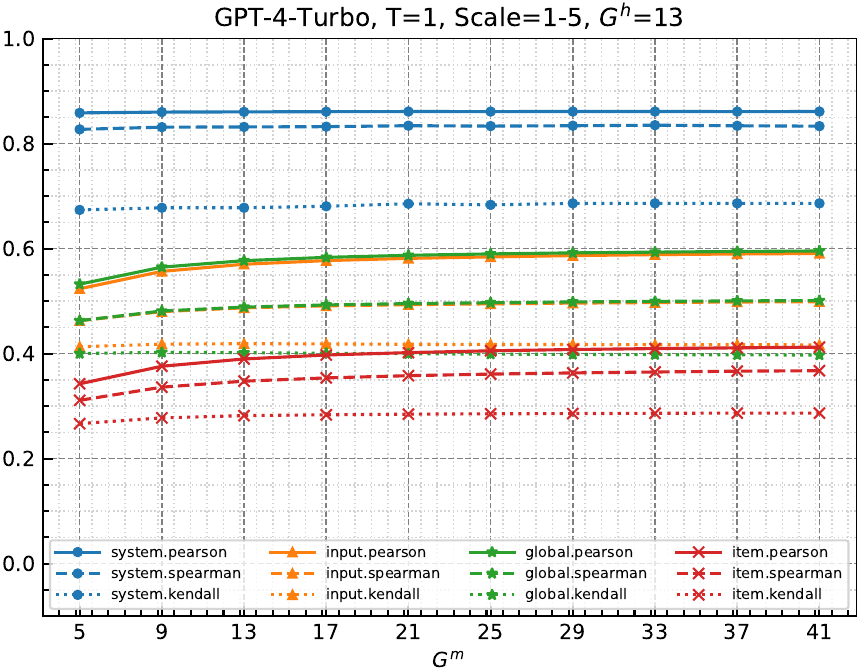} \hfill
  \includegraphics[width=0.49\linewidth]{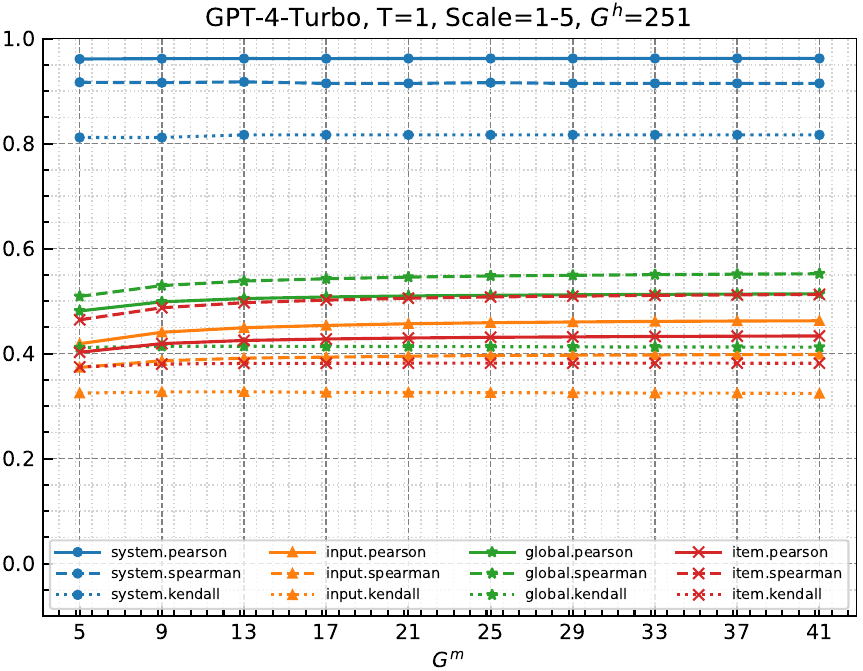}
  \caption{As the changes of $G^m$, the correlations between the GPT-4-Turbo evaluator and human evaluation using different measures on SummEval (left) and WMT23 (right) with the fixed evaluation scale of 1-5.}
  \label{fig:real}
\end{figure*}

\subsection{Sensitivity to Score Granularity}
\label{sec:granularity}

We first introduce the concepts of \textbf{evaluation scale} and \textbf{score granularity}, which apply to both human evaluation and emerging LLM-based evaluators. In practice, human evaluation scores are not continuous values like traditional metrics (e.g., BLEU, BERTScore). Evaluators typically select a discrete value as the evaluation score based on the scale (also called range) required by the evaluation guidelines (e.g., 1-5, 0-100). This scale is referred to as the \textbf{evaluation scale}. On the other hand, multiple evaluators are often involved to enhance the reliability of the evaluation, and the final human score is the average of the scores from multiple evaluators, leading to more diverse values. The combination of the chosen evaluation scale and the number of evaluators determines the number of all possible final scores, which is called \textbf{score granularity}. This reflects the degree of discretization in the evaluation and potential ties in the scores. For example, SummEval uses a 5-point Likert scale for human evaluation, and each sample is evaluated by three annotators, resulting in 13 possible averaged scores. Here, the evaluation scale is 1-5, and the score granularity is 13.

Moreover, LLMs with strong instruction-following capabilities like GPT-4 have been increasingly used in automatic evaluation, following the pattern of humans \citep{DBLP:conf/eamt/KocmiF23, DBLP:conf/acl/ChiangL23}. Therefore, their output scores are also discrete, and they can simulate multiple evaluators through repeated sampling, which also involves the evaluation scale and score granularity. We have presented statistics for these two quantities in human scores and LLM-based metric scores across different datasets in Tables \ref{tab:subdata_parameter} and \ref{tab:metric_parameter} in the appendix. The differences in evaluation scales primarily affect the difficulty of evaluation and the ability of the evaluators, which are not the focus of our study. For example, prompting GPT-4-Turbo with the scale of 1-5 or 1-10 is often seen as two distinct evaluation metrics that possess different capabilities. Our main focus, instead, is the impact of score granularity differences on different correlation measures when the evaluation scale and metric remain the same, which can be experimented on real-world data and statistical simulation. We denote the score granularity of humans and metrics as $G^h$ and $G^m$, respectively, with the former depending on the datasets and the latter depending on the evaluation settings.

\subsubsection{Real-world Data}

We select two typical datasets from Table \ref{tab:dataset_meta_info} for experiments: SummEval and WMT23, where their $G^h$ equals to 13 and 251, respectively. In addition, we use the three proprietary LLMs and two evaluation scales (1-5, 1-10) introduced in Section \ref{sec:metrics}, with the temperature setting of T=1. The number of multiple samplings varies from 1 to 10 to obtain results for different levels of metric score granularity $G^m$. Figure \ref{fig:real} shows the consistency of GPT-4-Turbo with human evaluations under different correlation measures, with other results included in Figures \ref{fig:real1}-\ref{fig:real10} in the appendix. We believe that when the score granularity of the metric $G^m$ is no greater than that of human $G^h$, the evaluation capability of the metric is limited, as fewer possible scores are available. Consequently, as $G^m$ increases until it equals $G^h$, the consistency between the metric and human should increase. However, when $G^m$ exceeds $G^h$, the situation becomes complex, and it is difficult to intuitively estimate the expected trend of the consistency, leaving for future explorations. The results illustrate that the measures using the system-level grouping or Kendall's $\tau$ are basically not affected by the score granularity, where the consistency almost remains unchanged, not meeting expectations as mentioned above. The situation is generally similar for the two datasets with different LLM-based evaluators and different evaluation scales.

\begin{figure}[t]
  \includegraphics[width=\linewidth]{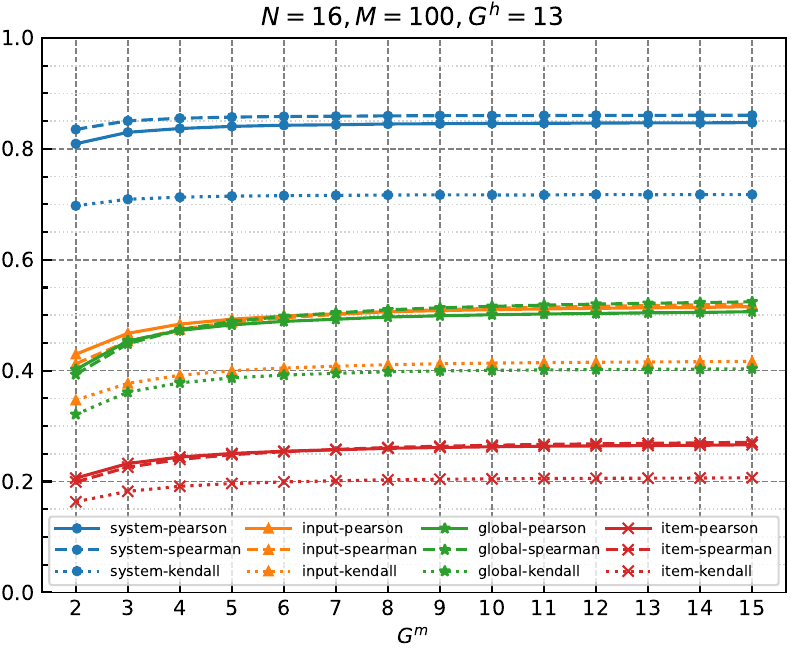}
  \caption{As the changes of $G^m$, the correlations between metrics and humans using different measures in statistical simulation with $G^h=13$.}
  \label{fig:simulation}
\end{figure}

\subsubsection{Statistical Simulation}

Although real-world datasets and evaluation metrics are more reliable for analysis compared to synthetic data, their scale of data is limited and some variables are less flexible to adjust. Therefore, we also employ the statistical simulation as an auxiliary method to observe whether similar conclusions can be drawn. Based on our collected datasets and certain hypotheses, we model the NLG meta-evaluation data and conduct experiments with parameters estimated from real data. The detailed process, including the sampling algorithm and experimental settings, is described in Appendix \ref{sec:simulation}. Figure \ref{fig:simulation}, as well as Figure \ref{fig:simulation1} in the appendix, show the results, which generally align with the observations obtained before.

\paragraph{Takeaways} The influence of the score granularity on the measures using system-level grouping or Kendall correlation is the least among different grouping methods and correlation coefficients, which is not as expected. And as the score granularity of the metric increases, the consistency between metrics and humans under other correlation measures continuously increases.

\section{Related Work}

In the field of NLP, there is limited research analyzing correlation measures in NLG meta-evaluation. \citet{DBLP:conf/acl/MathurBC20} found that the introduction of an outlier system can distort the system-level Pearson correlation between machine translation evaluation metrics and human evaluation and quantify the impact. \citet{DBLP:conf/naacl/DeutschDR22} proposed two different ways of calculating system-level correlations by changing the input sets or system pairs and showed they can lead to more precise estimates of metric performance in real-world scenarios. Recently, \citet{DBLP:conf/emnlp/DeutschFF23} pointed out that the segment-level Kendall correlation coefficient, widely used in machine translation evaluation, does not handle ties in human scores and metric outputs as expected and thus needs to be calibrated. It is worth mentioning that \citet{perrella-etal-2024-guardians} designed several sentinel metrics for machine translation meta-evaluation and showed that global-level grouping and system-level grouping may introduce unfairness.

In the research on automatic evaluation metrics, some works have commented on correlation measures based on experimental results when presenting the performance of different evaluation metrics. \citet{DBLP:conf/acl/OwczarzakRDC12} found that, in the domain of summarization evaluation, system-level correlation is more robust to inconsistent human annotations. \citet{DBLP:conf/wmt/FreitagRMLSAKFLM22} discovered that system-level correlation is hard to distinguish between different machine translation evaluation metrics. \citet{DBLP:conf/emnlp/LiuIXWXZ23,DBLP:conf/emnlp/XuWPSFWL23} explained some experimental results as the inappropriate handling of ties by the Kendall correlation when comparing the performance of different metrics. Additionally, \citet{DBLP:conf/acl/WeiJ20} demonstrated the output scores of automatic evaluation metrics have less variance than human scores at system level.

In contrast, we focus on the properties and capabilities of typical correlation measures from generic NLG evaluation perspectives, not limited to specific tasks and evaluation metrics.

\section{Conclusions}

We analyze and compare the characteristics and capabilities of 12 typical correlation measures in NLG meta-evaluation through three proposed perspectives. Based on various experiments with large-scale real-word data from six NLG datasets and 32 evaluation metrics, we find measures using global-level grouping and Pearson correlation have better overall meta-evaluation capability, while those using system-level and Kendall correlation show the opposite. And they have all been used on some common evaluation datasets. We hope that our work can deepen the understanding of correlation measures and draw more attention and emphasis on related research in the future.




\section*{Limitations}
We mainly analyzed the capabilities of correlation measures through empirical experiments without conducting theoretical analysis. Besides, although our experiments have covered many common and typical evaluation datasets to provide as general analyses as possible, it is impossible to encompass all tasks and evaluation aspects in NLG. Therefore, the conclusions we obtained about the meta-evaluation capability of different correlation measures may be limited to a certain context. Additionally, our work requires quite a few resources, including the API cost of using proprietary LLMs to annotate data and high-performance computation for conducting large-scale empirical evaluations and statistical simulations, which could be improved in the future.

\section*{Acknowledgments}
This work was supported by Beijing Science and Technology Program (Z231100007423011) and Key Laboratory of Science, Technology and Standard in Press Industry (Key Laboratory of Intelligent Press Media Technology). We appreciate the anonymous reviewers for their helpful comments. Xiaojun Wan is the corresponding author.

\bibliography{custom}

\appendix

\begin{table*}
\centering\small
\begin{tabular}{p{15cm}}
\toprule
\textbf{Prompts and Instructions} \\
\midrule
  
\#\#\#Instruction\#\#\# \\
Please act as an impartial and helpful evaluator for natural language generation (NLG), and the audience is an expert in the field. \\
Your task is to evaluate the quality of \{task\} strictly based on the given evaluation criterion. \\
Begin the evaluation by providing your analysis concisely and accurately, and then on the next line, start with "Rating:" followed by your rating on a Likert scale from \{scale\} (higher means better). \\
You MUST keep to the strict boundaries of the evaluation criterion and focus solely on the issues and errors involved; otherwise, you will be penalized. \\
Make sure you read and understand these instructions, as well as the following evaluation criterion and example content, carefully. \\
\\
\#\#\#Evaluation Criterion\#\#\# \\
\{aspect\} \\
\\
\#\#\#Example\#\#\# \\
\{source\_des\}: \\
\{source\} \\
\\
\{target\_des\}: \\
\{target\} \\
\\
\#\#\#Your Evaluation\#\#\# \\
\bottomrule
\end{tabular}
\caption{Prompts and instructions used for LLMs to evaluate and annotate NLG tasks.}
\label{tab:prompt}
\end{table*}

\section{Definitions of Pearson's \texorpdfstring{$r$}{r},  Spearman's \texorpdfstring{$\rho$}{rho}, and Kendall's \texorpdfstring{$\tau$}{tau}}
\label{app:correlation_coefficient}

Pearson correlation coefficient measures the linear relationship between two variables. The formula is:

$$
r(x,y) = \frac{\sum_{i=1}^N (x_i - \bar{x})(y_i - \bar{y})}{\sqrt{\sum_{i=1}^N (x_i - \bar{x})^2 \sum_{i=1}^N (y_i - \bar{y})^2}}
$$

Where $x$ and $y$ are n-dimensional vectors, and $x_i$ and $y_i$ are the elements of $x$ and $y$. $\bar{x}=\sum_{i=1}^n x_i$ and $\bar{y}=\sum_{i=1}^n y_i$.

Spearman correlation coefficient \citep{ca468a70-0be4-389a-b0b9-5dd1ff52b33f} measures the rank-order correlation between two variables. The formula is:

$$
\rho(x,y) = 1 - \frac{6 \sum_{i=1}^n d_i^2}{n(n^2 - 1)}
$$

Where $d_i = \text{rank}(x_i) - \text{rank}(y_i)$ is the difference between the ranks of corresponding values in the two vectors.

Kendall correlation coefficient \citep{10.1093/biomet/33.3.239} measures the ordinal association between two variables. It considers concordant and discordant pairs, with $\tau_b$ adjusted for ties. The formula is:

$$
\tau_b(x,y) = \frac{C - D}{\sqrt{(C + D + T_x)(C + D + T_y)}}
$$

Where $C$ is the number of concordant pairs, i.e. pairs of elements $(i,j)$ where $(x_i > x_j \land y_i > y_j) \lor (x_i < x_j \land y_i < y_j)$. $D$ is the number of discordant pairs, i.e. pairs of elements $(i,j)$ where $(x_i > x_j \land y_i < y_j) \lor (x_i < x_j \land y_i > y_j)$. $T_x$ and $T_y$ are the number of tied values in $x$ and $y$, respectively.

\section{Details of Selected Evaluation Metrics}
\label{app:evaluation_metrics}

\subsection{Non-LLM evaluation metrics}

For CHRF and BLEU, we use the implementation of TorchMetrics\footnote{\url{https://lightning.ai/docs/torchmetrics/stable/}}. For ROUGE, BERTSCORE, and BLEURT, we use the \texttt{evaluation} package of Huggingface with the default parameters. For MoverScore\footnote{\url{https://github.com/AIPHES/emnlp19-moverscore}}, BARTScore\footnote{\url{https://github.com/neulab/BARTScore}}, and COMET\footnote{\url{https://github.com/Unbabel/COMET}}, we use the code from the original GitHub repositories and the default models. We check the licenses of all open source programs to ensure that our use is compliant.

\subsection{Evaluation Prompts for LLMs}

We used the same prompts to instruct GPT-3.5, GPT-4, and GPT-4o for NLG evaluation. To save space, we present a template of our prompt in Table \ref{tab:prompt}. We filled the aspect part of the prompt with definitions from original datasets. When the original dataset lacked these definitions, we composed them based on our understanding.

\subsection{Reasons not to use non-proprietary LLMs}

We originally intended to include some open-source LLMs (e.g., Llama 3 \citep{dubey2024llama}). However, we found that their adherence to "evaluation instructions" was not strong enough, often resulting in non-compliant responses. We also considered including some fine-tuned LLM evaluators (e.g., Themis \citep{DBLP:journals/corr/abs-2406-18365}). However, their limitations due to fine-tuning settings prevented us from requiring them to score with different evaluation scales via prompts. Therefore, we finally only chose proprietary LLMs in our experiments.

\section{Process of Statistical Simulation}
\label{sec:simulation}

Based on the real-world data and some hypotheses, we first establish a probabilistic model for NLG meta-evaluation and then obtain results through the proposed algorithm and repeated sampling.

\subsection{Modeling NLG Meta-Evaluation}

We posit that the capability of an evaluation metric is primarily reflected in two aspects: the ability to evaluate the overall level of different systems and the ability to evaluate different output texts from a given system. In practice, system-level correlation can estimate the former, while item-level correlation can estimate the latter. Therefore, we treat these two quantities as the control parameters during modeling. We assume that the scores of the evaluation metric and human evaluation here are continuous, and for a system $s_i$, those of outputs it generates for various inputs follow a bivariate normal distribution: $x_{ij}, z_{ij} \sim \mathcal{N} (\mu^m_i, \mu^h_i, \sigma^m_i, \sigma^h_i, \rho_i)$, where $\rho_i$\footnote{This $\rho$ does not refer to the Spearman correlation coefficient, the same below.} controls the correlation between metrics and humans within a single system. Based on our observations and those of \citep{DBLP:conf/emnlp/ShenCNYB23}, this correlation varies across different systems for most evaluation metrics. For simplicity, we assume $\rho_i$ follows a truncated normal distribution: $\rho_i \sim \mathcal{N} (\mu_{\rho_{item}}, \sigma_{\rho_{item}})$. Since item-level correlation is defined as the mean correlation coefficient across different systems, its value in existing datasets can be viewed as an estimate of $\mu_{\rho_{item}}$. Furthermore, assuming $\mu^m_i $ and $\mu^h_i $ of the above bivariate normal distribution follow another bivariate normal distribution: $\mu^m_i, \mu^h_i \sim \mathcal{N} (\mu^m, \mu^h, \sigma^m, \sigma^h, \rho_{sys})$, where $\rho_{sys} $ controls the correlation between $\mu^m_i$ and $\mu^h_i$. And similarly, system-level correlation in real-world scenario can be seen as an estimate of $\rho_{sys}$ because $\frac{1}{M}\sum_{j=1}^{M}x_{ij}$ and $\frac{1}{M}\sum_{j=1}^{M}z_{ij}$ in the definition of system-level correlation are viewed as estimates of $\mu^m_i$ and $\mu^h_i$.

Although the assumption of continuous scores makes it convenient to model and sample data, we need to analyze the impact of score granularity, which requires further data discretization. We then follow the practice of \citet{onoshima2019decline} that has a similar situation to sample $G^h-1$ and $G^m-1$ thresholds from uniform distributions $U(\mu^m-\sigma^m, \mu^m+\sigma^m)$ and $U(\mu^h-\sigma^h, \mu^h+\sigma^h)$ to discretize them. The Algorithm \ref{alg:simulation} shows the pseudocode for the entire sampling process.

\begin{algorithm}[H]
{\small
\caption{Statistical Simulation}
\label{alg:simulation}
 \textbf{Input:} $\mu^m,\mu^h,\sigma^m,\sigma^h,\sigma^m_i,\dots,\sigma^m_N,\sigma^h_1,\dots,\sigma^h_N \in \mathbb{R}, N,$ 
 $M,G^m,G^h,T_1, T_2 \in \mathbb{N},$ $\rho_{sys},\mu_{\rho_{item}},\sigma_{\rho_{item}}, C$.

\textbf{Output:} Correlation coefficient
\begin{algorithmic}[0]
\State $R \gets$ an empty list
\For{$T_1$ iterations}
    \State $X^s, Z^s \gets$ empty $N \times M$ matrices
    \For{$i \in \{1, \dots, N\}$}
        \State sample $\mu^m_i,\mu^h_i \sim \mathcal{N}(\mu^m, \mu^h, \sigma^m, \sigma^h, \rho_{sys})$
        \State sample $\rho_i \sim \mathcal{N}(\mu_{\rho_{item}}, \sigma_{\rho_{item}})$
        \For{$j \in \{1, \dots, M\}$}
            \State sample $x_{ij}, z_{ij} \sim \mathcal{N} (\mu^m_i, \mu^h_i, \sigma^m_i, \sigma^h_i, \rho_i)$
            \State $X^s[i, j] \gets x_{ij}$
            \State $Z^s[i, j] \gets z_{ij}$
        \EndFor
    \EndFor
    \For{$T_2$ iterations}
        \State sample $\{t^m_n\}_{n=1}^{G^m-1} \sim U(\mu^m-\sigma^m, \mu^m+\sigma^m)$
        \State sample $\{t^h_n\}_{n=1}^{G^h-1} \sim U(\mu^h-\sigma^h, \mu^h+\sigma^h)$
        \State $X^s \gets$ \Call{Discretize}{$X^s, \{t^m_n\}_{n=1}^{G^m-1}$}
        \State $Z^s \gets$ \Call{Discretize}{$Z^s, \{t^h_n\}_{n=1}^{G^h-1}$}
        \State $C^s \gets C(X^s, Z^s)$
        \State Add $C^s$ to $R$  
    \EndFor
\EndFor
\State \Return \Call{Avg}{$R$}
\end{algorithmic}

\begin{algorithmic}[0]
\Function{Discretize}{$X, \{t_n\}_{n=1}^{G}$}
\State $Y \gets$ zero $N \times M$ matrix
\For{$(i, j) \in \{1, \dots, N\} \times \{1, \dots, M\}$}
    \For{$k \in \{1, \dots, G\}$}
        \If{$X[i, j]\leq t_k$}
            \State $Y[i, j] \gets k$
            \State \text{BREAK}
        \EndIf
    \EndFor
    \If{$Y[i, j]$ is 0}
        \State $Y[i, j] \gets G+1$
    \EndIf
\EndFor
\State \Return $Y$
\EndFunction
\end{algorithmic}
}
\end{algorithm}

\subsection{Experimental Settings}

For the data collected in Section \ref{sec:data_preparation}, all evaluation scores from humans and metrics are normalized to the 0-1 scale for parameter estimation, with results shown in Tables \ref{tab:subdata_parameter} and \ref{tab:metric_parameter}. We used the data where GPT-4-turbo is used as the evaluation metrics to estimate the parameters and the values are averaged across different meta-evaluation sets: $\mu^m = \mu^m_1 = \dots = \mu^m_N = 0.47, \mu^h = \mu^h_1 = \dots = \mu^h_N = 0.65$, $\sigma^m =\sigma^m_1 = \dots = \sigma^m_N = 0.16, \sigma^h = \sigma^h_1 = \dots = \sigma^h_N = 0.14$, $\sigma_{\rho_{item}} = 0.14$, $\rho_{sys} = 0.92$ and $\mu_{\rho_{item}} = 0.35$. For the number of systems and input documents, we consider $N = 16, M = 100$, which is typical and the same as SummEval. The corresponding $G^h$ is set to two common values: 13 and 250, and $G^m$ is selected from 2 to 15 and 5 to 100, respectively. Due to the huge amount of computation, we set $T_1=T_2=100$.

\section{Other Figures and Tables}
\label{sec:appendix}
Given the limited space of the main text, we present the complete experimental results here.

\begin{figure}[!htp]
  \includegraphics[width=1\linewidth]{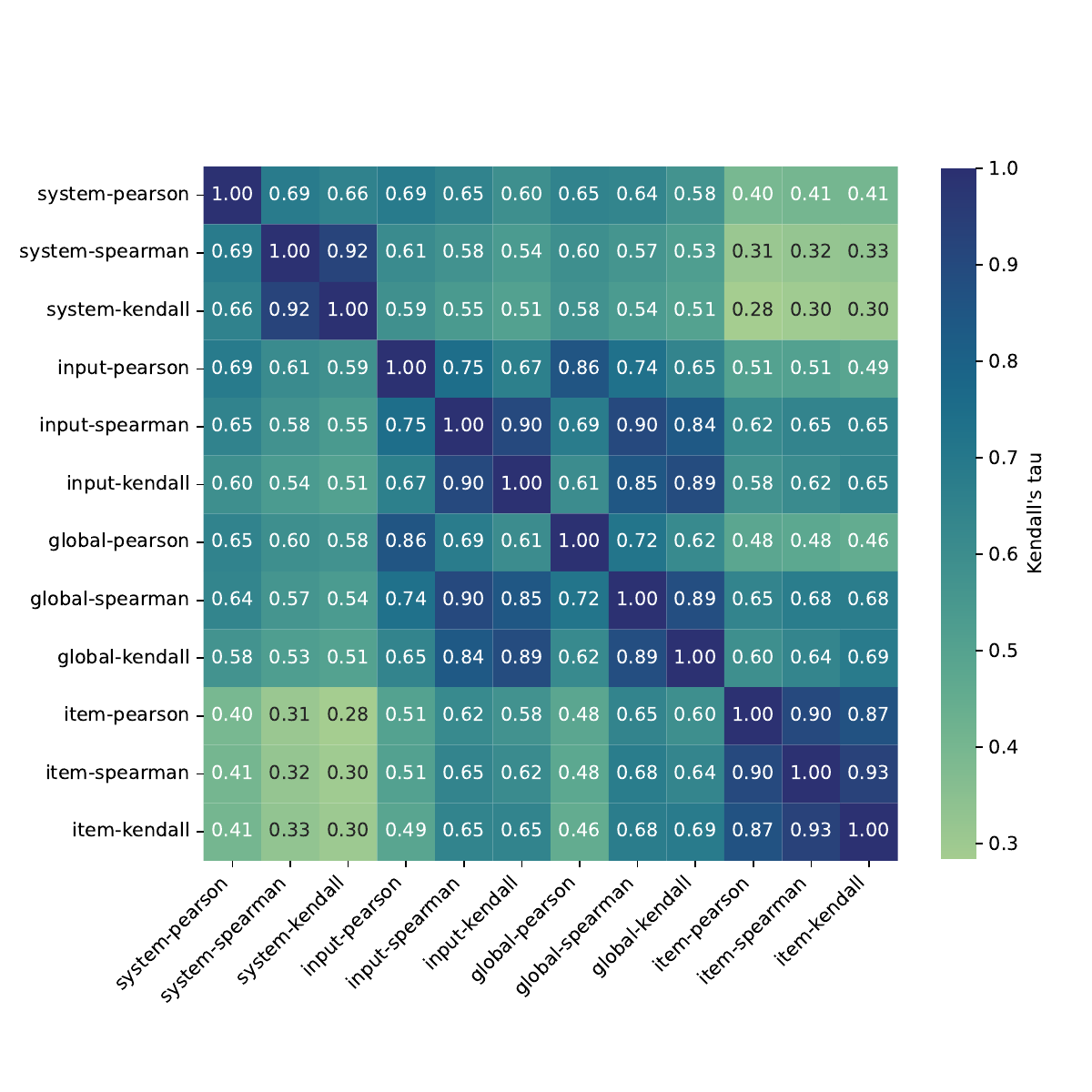}
  \caption{The consistency of evaluation metric rankings using different correlation measures on HANNA, calculated through Kendall's correlation coefficient.}
  \label{fig:hanna}
\end{figure}

\begin{figure}[!htp]
  \includegraphics[width=1\linewidth]{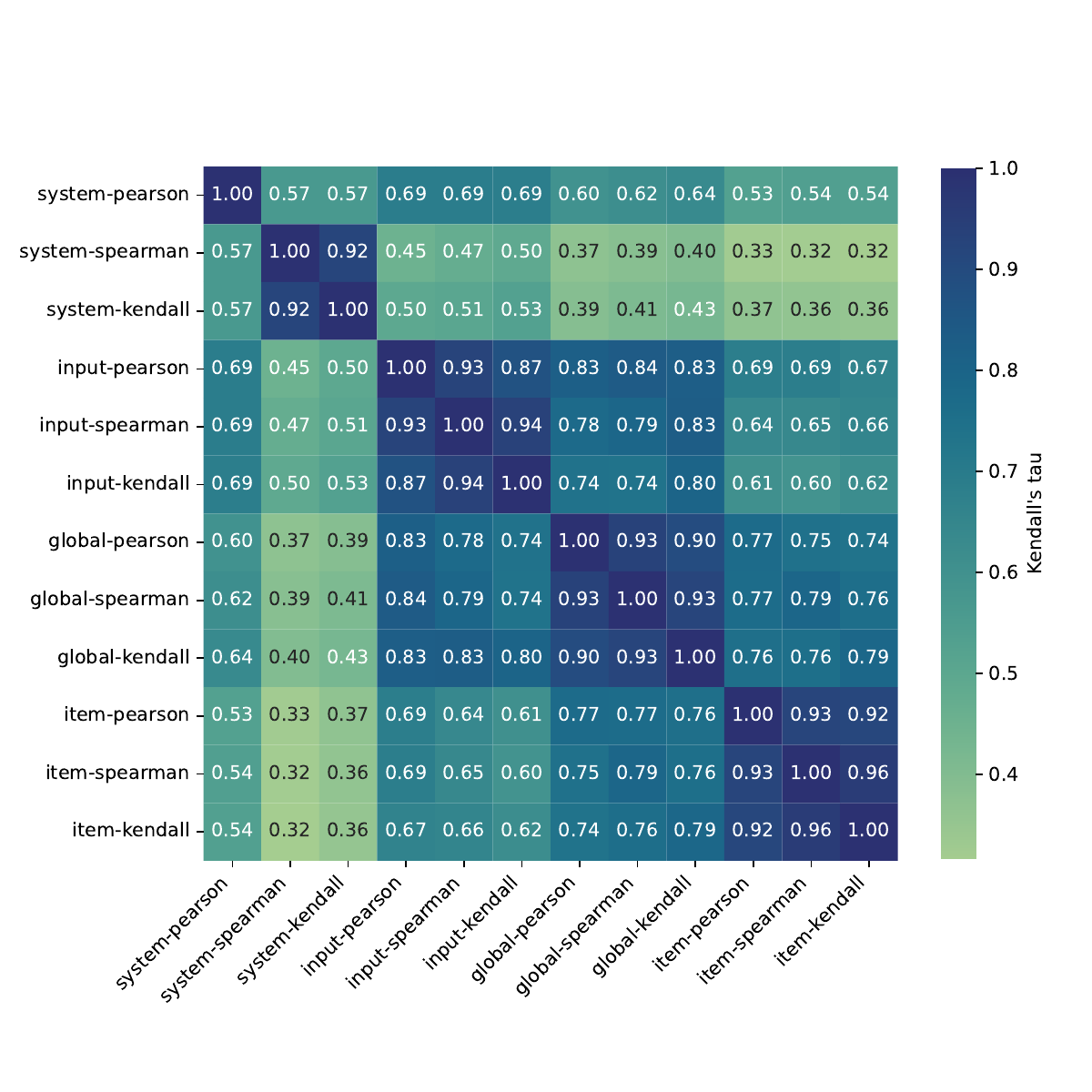}
  \caption{The consistency of evaluation metric rankings using different correlation measures on MANS, calculated through Kendall's correlation coefficient.}
  \label{fig:mans}
\end{figure}

\begin{figure}[!htp]
  \includegraphics[width=1\linewidth]{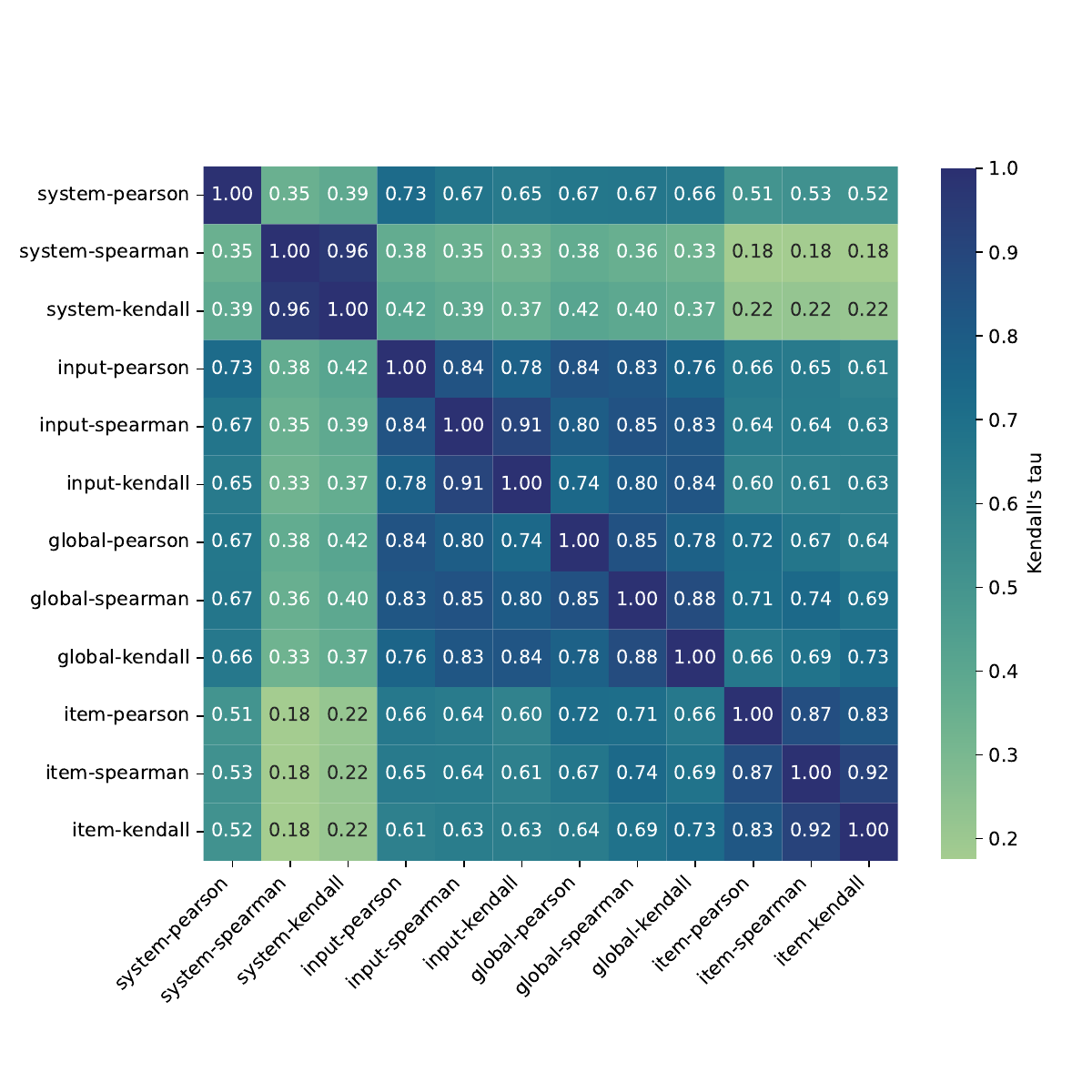}
  \caption{The consistency of evaluation metric rankings using different correlation measures on USR, calculated through Kendall's correlation coefficient.}
  \label{fig:usr}
\end{figure}

\begin{figure}[!htp]
  \includegraphics[width=1\linewidth]{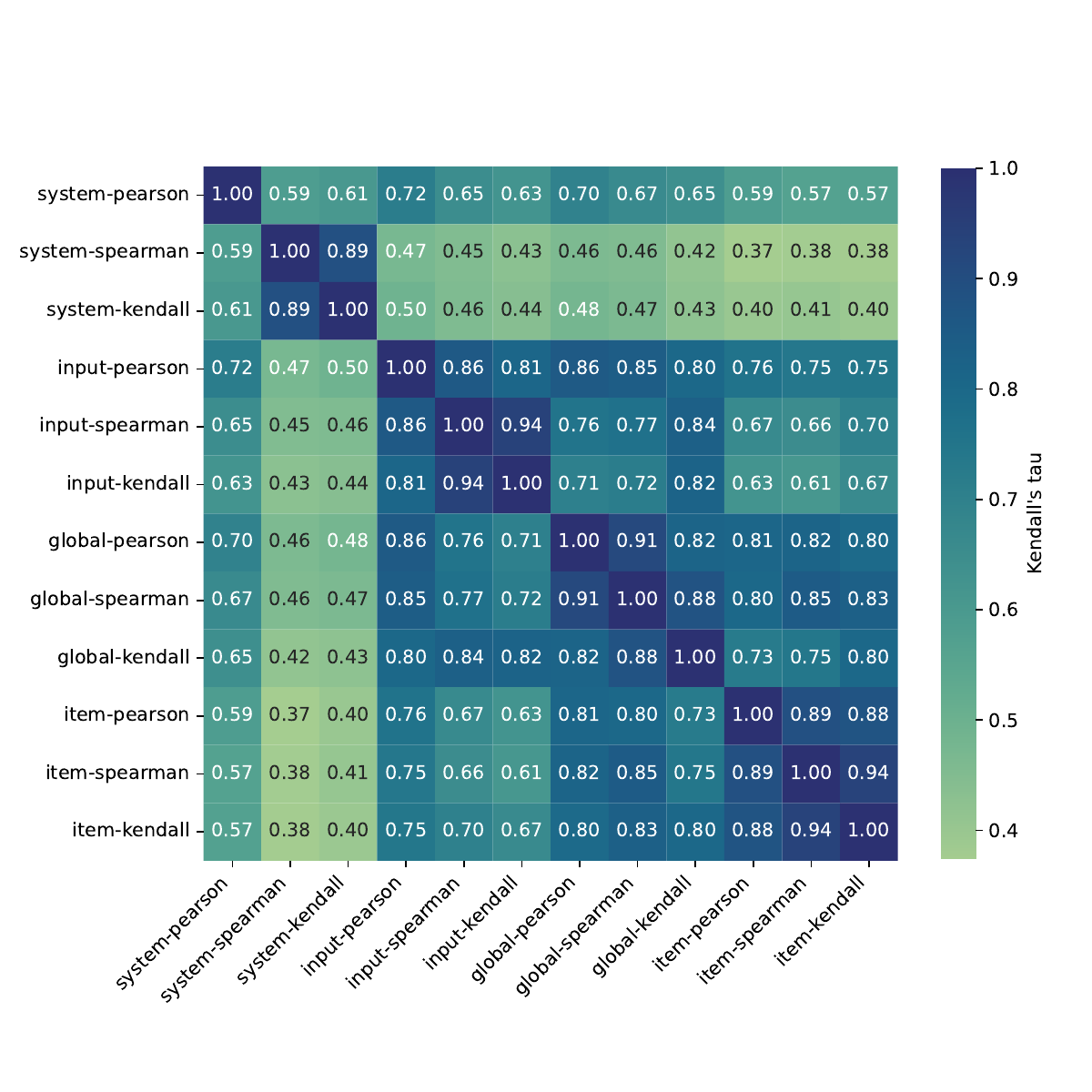}
  \caption{The consistency of evaluation metric rankings using different correlation measures on WebNLG2020, calculated through Kendall's correlation coefficient.}
  \label{fig:webnlg}
\end{figure}

\begin{figure}[!htp]
  \includegraphics[width=1\linewidth]{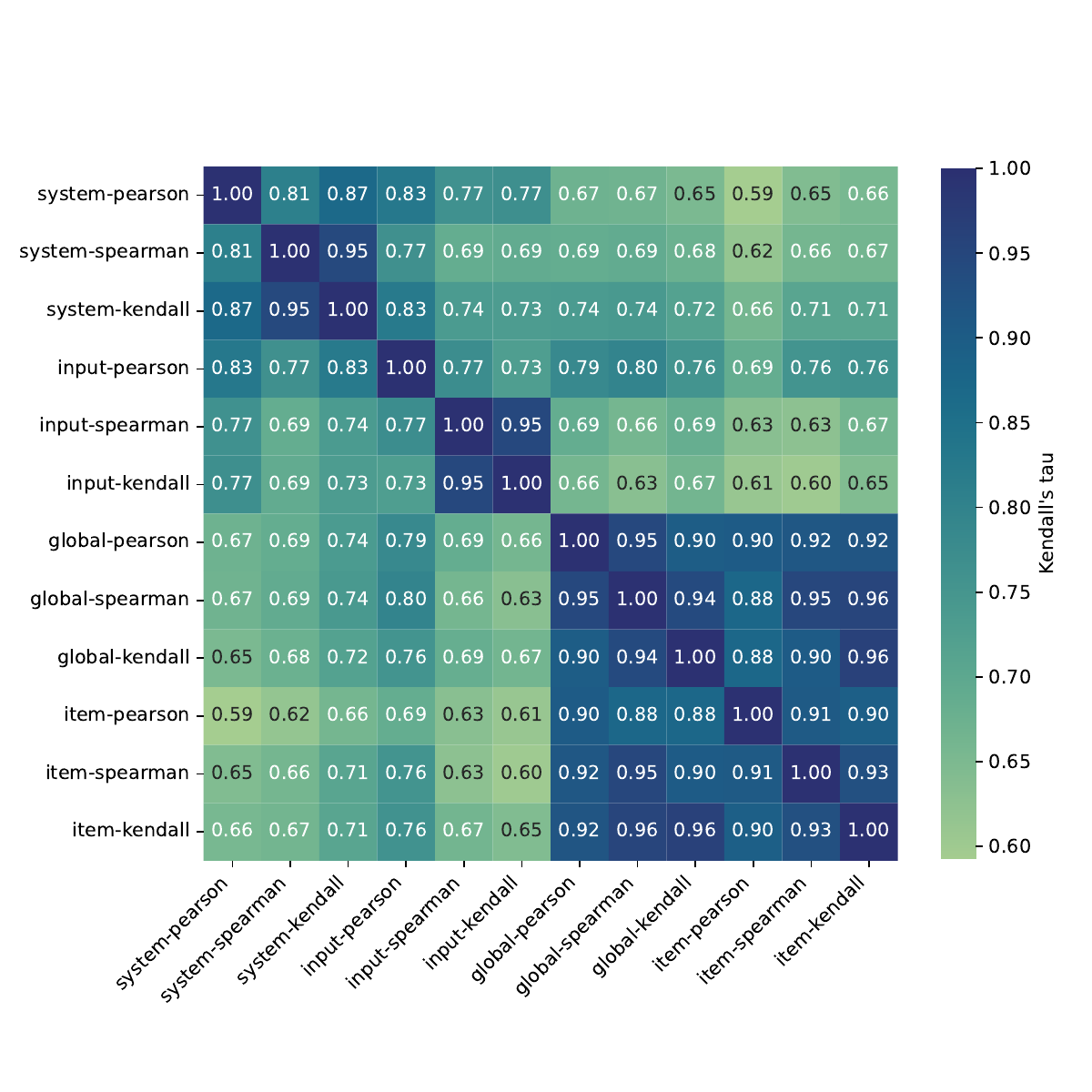}
  \caption{The consistency of evaluation metric rankings using different correlation measures on WMT23-ZH2EN-NEWS, calculated through Kendall's correlation coefficient.}
  \label{fig:wmt}
\end{figure}

\newpage
\begin{table*}[]
\small
\centering
\begin{tabular}{llllccccc}
\hline
Dataset & Subset & Aspect & Label & $\widehat{\mu^h}$ & $\widehat{\sigma^h}$ & $G^h$ & Scale & Tie Ratio \\ \hline
SummEval & CNN/DM & Coherence & D1 & 0.60 & 0.15 & 13 & 1-5 & 0.10 \\
SummEval & CNN/DM & Consistency & D2 & 0.92 & 0.14 & 13 & 1-5 & 0.67 \\
SummEval & CNN/DM & Fluency & D3 & 0.92 & 0.09 & 13 & 1-5 & 0.53 \\
SummEval & CNN/DM & Relevance & D4 & 0.69 & 0.09 & 13 & 1-5 & 0.13 \\
WMT23 & GeneralMT2023\_NEWS & Overall Quality & D5 & 0.84 & 0.05 & 251 & 0-250 & 0.11 \\
HANNA & WP & Coherence & D6 & 0.54 & 0.13 & 13 & 1-5 & 0.13 \\
HANNA & WP & Complexity & D7 & 0.36 & 0.14 & 13 & 1-5 & 0.12 \\
HANNA & WP & Empathy & D8 & 0.32 & 0.09 & 13 & 1-5 & 0.13 \\
HANNA & WP & Engagement & D9 & 0.42 & 0.13 & 13 & 1-5 & 0.12 \\
HANNA & WP & Relevance & D10 & 0.41 & 0.14 & 13 & 1-5 & 0.10 \\
HANNA & WP & Surprise & D11 & 0.28 & 0.10 & 13 & 1-5 & 0.15 \\
MANS& ROC & Overall & D12 & 0.38 & 0.15 & 21 & 1-3 & 0.06 \\
MANS& WP & Overall & D13 & 0.45 & 0.08 & 21 & 1-3 & 0.08 \\
USR & Persona-Chat & Engaging & D14 & 0.76 & 0.13 & 7 & 1-3 & 0.24  \\
USR & Persona-Chat & Maintains Context & D15 & 0.74 & 0.19 & 7 & 1-3 & 0.31  \\
USR & Persona-Chat & Natural & D16 & 0.89 & 0.08 & 7 & 1-3 & 0.48 \\
USR & Persona-Chat & Overall & D17 & 0.69 & 0.19 & 13 & 1-5 & 0.11  \\
USR & Persona-Chat & Understandable & D18 & 0.96 & 0.04 & 4 & 0-1 & 0.84  \\
USR & Persona-Chat & Uses Knowledge & D19 & 0.45 & 0.35 & 4 & 0-1 & 0.38  \\
USR & Topical-Chat & Engaging & D20 & 0.57 & 0.25 & 7 & 1-3 & 0.15  \\
USR & Topical-Chat & Maintains Context & D21 & 0.62 & 0.23 & 7 & 1-3 & 0.17 \\
USR & Topical-Chat & Natural & D22 & 0.64 & 0.21 & 7 & 1-3 & 0.17  \\
USR & Topical-Chat & Overall & D23 & 0.54 & 0.27 & 13 & 1-5 & 0.08  \\
USR & Topical-Chat & Understandable & D24 & 0.67 & 0.23 & 4 & 0-1 & 0.33  \\
USR & Topical-Chat & Uses Knowledge & D25 & 0.55 & 0.24 & 4 & 0-1 & 0.33  \\
WebNLG2020 & WebNLG2020 & Correctness & D26 & 0.88 & 0.07 & 401 & 0-100 & 0.03 \\
WebNLG2020 & WebNLG2020 & Data Coverage & D27 & 0.9 & 0.06 & 401 & 0-100 & 0.04 \\
WebNLG2020 & WebNLG2020 & Fluency & D28 & 0.83 & 0.06 & 401 & 0-100 & 0.01 \\
WebNLG2020 & WebNLG2020 & Relevance & D29 & 0.91 & 0.05 & 401 & 0-100 & 0.05 \\
WebNLG2020 & WebNLG2020 & Text Structure & D30 & 0.87 & 0.05 & 401 & 0-100 & 0.02 \\
\hline
\end{tabular}
\caption{Information of all the meta-evaluation datasets and estimated parameters.}
\label{tab:subdata_parameter}
\end{table*}

\begin{table*}[]
\centering
\small
\begin{tabular}{lcccccccccc}
\hline
Metric & $\widehat{\mu^m}$ & $\widehat{\sigma^m}$ & $\widehat{\rho_{sys}}$ & $\widehat{\mu_{\rho_{item}}}$ & $\widehat{\sigma_{\rho_{item}}}$ & $r_{\widetilde{N}}$ & $r_{N\times M}$ & $G^m$ & Scale & Tie Ratio \\ \hline
GPT3.5\_T=0\_0\_100 & 0.44 & 0.08 & 0.88 & 0.22 & 0.13 & 0.41 & 0.38 & 101 & 0-100 & 0.35 \\
GPT3.5\_T=0\_1\_10 & 0.33 & 0.08 & 0.83 & 0.20 & 0.13 & 0.41 & 0.35 & 10 & 1-10 & 0.46 \\
GPT3.5\_T=0\_1\_5 & 0.32 & 0.10 & 0.86 & 0.21 & 0.12 & 0.40 & 0.35 & 5 & 1-5 & 0.49 \\
GPT3.5\_T=1\_0\_100 & 0.45 & 0.08 & 0.89 & 0.27 & 0.14 & 0.48 & 0.44 & 1001 & 0-100 & 0.03 \\
GPT3.5\_T=1\_1\_10 & 0.33 & 0.08 & 0.88 & 0.26 & 0.13 & 0.46 & 0.41 & 91 & 1-10 & 0.10 \\
GPT3.5\_T=1\_1\_5 & 0.32 & 0.09 & 0.88 & 0.25 & 0.13 & 0.46 & 0.41 & 41 & 1-5 & 0.14 \\
GPT4\_T=0\_0\_100 & 0.51 & 0.15 & 0.92 & 0.35 & 0.14 & 0.56 & 0.55 & 101 & 0-100 & 0.22 \\
GPT4\_T=0\_1\_10 & 0.45 & 0.15 & 0.92 & 0.34 & 0.13 & 0.55 & 0.53 & 10 & 1-10 & 0.28 \\
GPT4\_T=0\_1\_5 & 0.44 & 0.18 & 0.92 & 0.31 & 0.14 & 0.55 & 0.51 & 5 & 1-5 & 0.45 \\
GPT4\_T=1\_0\_100 & 0.51 & 0.14 & 0.93 & 0.39 & 0.14 & 0.59 & 0.58 & 1001 & 0-100 & 0.02 \\
GPT4\_T=1\_1\_10 & 0.45 & 0.15 & 0.93 & 0.38 & 0.13 & 0.59 & 0.56 & 91 & 1-10 & 0.06 \\
GPT4\_T=1\_1\_5 & 0.43 & 0.17 & 0.92 & 0.36 & 0.14 & 0.58 & 0.55 & 41 & 1-5 & 0.18 \\
GPT4o\_T=0\_0\_100 & 0.48 & 0.14 & 0.92 & 0.37 & 0.13 & 0.57 & 0.55 & 101 & 0-100 & 0.21 \\
GPT4o\_T=0\_1\_10 & 0.42 & 0.15 & 0.91 & 0.35 & 0.12 & 0.55 & 0.53 & 10 & 1-10 & 0.27 \\
GPT4o\_T=0\_1\_5 & 0.40 & 0.17 & 0.91 & 0.32 & 0.13 & 0.54 & 0.50& 5 & 1-5 & 0.43 \\
GPT4o\_T=1\_0\_100 & 0.48 & 0.14 & 0.92 & 0.39 & 0.13 & 0.59 & 0.58 & 1001 & 0-100 & 0.02 \\
GPT4o\_T=1\_1\_10 & 0.42 & 0.14 & 0.92 & 0.38 & 0.13 & 0.59 & 0.56 & 91 & 1-10 & 0.05 \\
GPT4o\_T=1\_1\_5 & 0.40 & 0.16 & 0.91 & 0.35 & 0.13 & 0.57 & 0.54 & 41 & 1-5 & 0.16 \\
BERTScore-f & 0.87 & 0.04 & 0.60 & -0.02 & 0.10 & 0.26 & 0.22 & / & / & 0.01 \\
BERTScore-p & 0.87 & 0.04 & 0.53 & -0.02 & 0.10 & 0.23 & 0.20 & / & / & 0.01 \\
BERTScore-r & 0.88 & 0.04 & 0.67 & -0.01 & 0.11 & 0.29 & 0.24 & / & / & 0.01 \\
BLEU & 0.15 & 0.24 & 0.52 & -0.01 & 0.09 & 0.24 & 0.21 & / & / & 0.53 \\
CHRF & 0.37 & 0.20 & 0.58 & 0.00 & 0.10 & 0.27 & 0.24 & / & / & 0.02 \\
COMET & 0.60 & 0.12 & 0.73 & -0.01 & 0.11 & 0.32 & 0.27 & / & / & 0.00 \\
MoverScore & 0.61 & 0.11 & 0.54 & -0.02 & 0.11 & 0.24 & 0.22 & / & / & 0.00 \\
ROUGE-1 & 0.41 & 0.20 & 0.53 & -0.02 & 0.11 & 0.25 & 0.22 & / & / & 0.02 \\
ROUGE-2 & 0.23 & 0.23 & 0.54 & -0.01 & 0.10 & 0.25 & 0.22 & / & / & 0.15 \\
ROUGE-L & 0.33 & 0.21 & 0.53 & -0.03 & 0.10 & 0.23 & 0.20 & / & / & 0.02 \\ \hline
\end{tabular}
\caption{Metrics information and estimated parameters averaged across meta-evaluation datasets. We do not use the output of BLEURT and BARTScore-(s-h, r-h, h-r) to estimate the parameters because their scores do not have clear ranges.}
\label{tab:metric_parameter}
\end{table*}

\begin{table*}[!htp]
\small
\resizebox{\linewidth}{!}{
\begin{tabular}{llllllllllll}
\hline
Grouping & Correlation & D1 & D2 & D3 & D4 & D5 & D6 & D7 & D8 & D9 & D10 \\ \hline
System & Pearson & 0.074 (5) & 0.054 (2) & 0.148 (10) & 0.150 (10) & 0.056 (7) & 0.092 (3) & 0.035 (1) & 0.065 (1) & 0.055 (1) & 0.125 (7) \\
System & Spearman & 0.137 (9) & 0.141 (11) & 0.135 (7) & 0.174 (11) & 0.159 (11) & 0.195 (11) & 0.174 (9) & 0.213 (11) & 0.175 (8) & 0.209 (11) \\
System & Kendall & 0.178 (12) & 0.138 (10) & 0.172 (12) & 0.228 (12) & 0.171 (12) & 0.251 (12) & 0.214 (12) & 0.263 (12) & 0.219 (12) & 0.246 (12) \\
Input & Pearson & 0.053 (1) & 0.064 (3) & 0.059 (2) & 0.082 (2) & 0.077 (9) & 0.106 (6) & 0.085 (5) & 0.144 (7) & 0.093 (3) & 0.127 (8) \\
Input & Spearman & 0.074 (4) & 0.086 (6) & 0.096 (5) & 0.103 (6) & 0.085 (10) & 0.117 (7) & 0.097 (6) & 0.121 (5) & 0.117 (7) & 0.096 (2) \\
Input & Kendall & 0.074 (6) & 0.070 (4) & 0.099 (6) & 0.107 (8) & 0.061 (8) & 0.105 (5) & 0.101 (7) & 0.122 (6) & 0.116 (6) & 0.107 (4) \\
Global & Pearson & 0.064 (2) & 0.054 (1) & 0.058 (1) & 0.079 (1) & 0.035 (3) & 0.089 (2) & 0.070 (3) & 0.121 (4) & 0.085 (2) & 0.084 (1) \\
Global & Spearman & 0.071 (3) & 0.114 (8) & 0.090 (3) & 0.100 (5) & 0.023 (1) & 0.095 (4) & 0.067 (2) & 0.095 (2) & 0.100 (5) & 0.100 (3) \\
Global & Kendall & 0.085 (7) & 0.084 (5) & 0.096 (4) & 0.097 (4) & 0.053 (6) & 0.084 (1) & 0.079 (4) & 0.115 (3) & 0.093 (4) & 0.117 (5) \\
Item & Pearson & 0.135 (8) & 0.102 (7) & 0.146 (9) & 0.094 (3) & 0.038 (4) & 0.138 (8) & 0.162 (8) & 0.175 (8) & 0.175 (9) & 0.122 (6) \\
Item & Spearman & 0.147 (10) & 0.134 (9) & 0.149 (11) & 0.106 (7) & 0.027 (2) & 0.141 (9) & 0.180 (10) & 0.177 (9) & 0.179 (11) & 0.132 (9) \\
Item & Kendall & 0.151 (11) & 0.147 (12) & 0.145 (8) & 0.108 (9) & 0.041 (5) & 0.147 (10) & 0.206 (11) & 0.190 (10) & 0.179 (10) & 0.148 (10) \\ \hline
Grouping & Correlation & D11 & D12 & D13 & D14 & D15 & D16 & D17 & D18 & D19 & D20 \\ \hline
System & Pearson & 0.062 (1) & 0.214 (10) & 0.170 (7) & 0.171 (5) & 0.162 (10) & 0.147 (4) & 0.244 (10) & 0.451 (10) & 0.209 (1) & 0.101 (4) \\
System & Spearman & 0.188 (8) & 0.433 (11) & 0.424 (11) & 0.374 (11) & 0.355 (11) & 0.359 (11) & 0.375 (11) & 0.544 (11) & 0.349 (3) & 0.579 (11) \\
System & Kendall & 0.225 (12) & 0.444 (12) & 0.505 (12) & 0.386 (12) & 0.358 (12) & 0.374 (12) & 0.396 (12) & 0.623 (12) & 0.361 (4) & 0.651 (12) \\
Input & Pearson & 0.135 (5) & 0.102 (4) & 0.135 (4) & 0.167 (4) & 0.119 (3) & 0.170 (5) & 0.116 (3) & 0.351 (7) & 0.336 (2) & 0.092 (2) \\
Input & Spearman & 0.123 (4) & 0.114 (6) & 0.141 (5) & 0.182 (6) & 0.145 (7) & 0.186 (6) & 0.158 (7) & 0.392 (9) & 0.414 (5) & 0.138 (6) \\
Input & Kendall & 0.151 (7) & 0.106 (5) & 0.145 (6) & 0.189 (7) & 0.157 (8) & 0.194 (9) & 0.164 (8) & 0.373 (8) & 0.414 (6) & 0.143 (7) \\
Global & Pearson & 0.136 (6) & 0.072 (1) & 0.107 (1) & 0.093 (1) & 0.127 (5) & 0.108 (1) & 0.088 (1) & 0.165 (1) & 0.540 (9) & 0.080 (1) \\
Global & Spearman & 0.105 (2) & 0.072 (2) & 0.116 (3) & 0.134 (3) & 0.094 (1) & 0.121 (3) & 0.092 (2) & 0.219 (3) & 0.487 (8) & 0.101 (3) \\
Global & Kendall & 0.107 (3) & 0.095 (3) & 0.110 (2) & 0.129 (2) & 0.110 (2) & 0.119 (2) & 0.128 (4) & 0.193 (2) & 0.483 (7) & 0.104 (5) \\
Item & Pearson & 0.201 (9) & 0.148 (7) & 0.192 (9) & 0.215 (8) & 0.127 (4) & 0.192 (8) & 0.148 (5) & 0.232 (4) & 0.569 (10) & 0.192 (8) \\
Item & Spearman & 0.214 (10) & 0.152 (8) & 0.186 (8) & 0.224 (9) & 0.140 (6) & 0.192 (7) & 0.152 (6) & 0.265 (6) & 0.603 (12) & 0.227 (9) \\
Item & Kendall & 0.219 (11) & 0.182 (9) & 0.196 (10) & 0.234 (10) & 0.157 (9) & 0.210 (10) & 0.180 (9) & 0.262 (5) & 0.601 (11) & 0.232 (10) \\ \hline
Grouping & Correlation & D21 & D22 & D23 & D24 & D25 & D26 & D27 & D28 & D29 & D30 \\ \hline
System & Pearson & 0.157 (7) & 0.140 (6) & 0.134 (6) & 0.149 (7) & 0.159 (1) & 0.092 (3) & 0.076 (2) & 0.155 (10) & 0.109 (3) & 0.127 (10) \\
System & Spearman & 0.563 (11) & 0.616 (11) & 0.642 (11) & 0.573 (11) & 0.602 (11) & 0.262 (11) & 0.155 (10) & 0.314 (11) & 0.472 (11) & 0.282 (11) \\
System & Kendall & 0.595 (12) & 0.669 (12) & 0.711 (12) & 0.622 (12) & 0.698 (12) & 0.290 (12) & 0.173 (12) & 0.408 (12) & 0.532 (12) & 0.337 (12) \\
Input & Pearson & 0.117 (4) & 0.110 (3) & 0.106 (3) & 0.107 (1) & 0.263 (6) & 0.110 (6) & 0.115 (3) & 0.080 (3) & 0.123 (4) & 0.092 (5) \\
Input & Spearman & 0.158 (8) & 0.136 (4) & 0.123 (5) & 0.147 (6) & 0.271 (7) & 0.138 (10) & 0.145 (9) & 0.101 (8) & 0.184 (10) & 0.125 (9) \\
Input & Kendall & 0.167 (10) & 0.140 (5) & 0.135 (7) & 0.145 (5) & 0.259 (5) & 0.128 (8) & 0.129 (6) & 0.084 (5) & 0.174 (9) & 0.106 (8) \\
Global & Pearson & 0.117 (3) & 0.096 (1) & 0.085 (1) & 0.114 (2) & 0.225 (3) & 0.064 (1) & 0.064 (1) & 0.054 (1) & 0.074 (1) & 0.061 (2) \\
Global & Spearman & 0.106 (2) & 0.104 (2) & 0.104 (2) & 0.117 (3) & 0.233 (4) & 0.088 (2) & 0.120 (5) & 0.068 (2) & 0.135 (5) & 0.058 (1) \\
Global & Kendall & 0.096 (1) & 0.153 (7) & 0.109 (4) & 0.118 (4) & 0.221 (2) & 0.107 (4) & 0.140 (8) & 0.093 (7) & 0.155 (6) & 0.077 (3) \\
Item & Pearson & 0.129 (5) & 0.168 (8) & 0.178 (9) & 0.193 (10) & 0.332 (8) & 0.127 (7) & 0.118 (4) & 0.083 (4) & 0.108 (2) & 0.095 (6) \\
Item & Spearman & 0.164 (9) & 0.197 (9) & 0.178 (8) & 0.180 (8) & 0.364 (9) & 0.108 (5) & 0.134 (7) & 0.091 (6) & 0.171 (7) & 0.087 (4) \\
Item & Kendall & 0.157 (6) & 0.247 (10) & 0.182 (10) & 0.191 (9) & 0.367 (10) & 0.135 (9) & 0.163 (11) & 0.110 (9) & 0.174 (8) & 0.104 (7) \\ \hline
\end{tabular}
}
\caption{The complete DP values of different correlation measures on all meta-evaluation datasets using permutation test, the lower the better, which are visualized as Figure \ref{fig:DP}.} 
\label{tab:DP}
\end{table*}

\begin{table*}[htp]
\small
\resizebox{\linewidth}{!}{
\begin{tabular}{llllllllllll}
\hline
Grouping & Correlation & D1 & D2 & D3 & D4 & D5 & D6 & D7 & D8 & D9 & D10 \\ \hline
System & Pearson & 0.844 (4) & 0.878 (2) & 0.673 (12) & 0.637 (10) & 0.853 (8) & 0.777 (7) & 0.928 (1) & 0.825 (1) & 0.876 (1) & 0.643 (10) \\
System & Spearman & 0.695 (8) & 0.756 (10) & 0.724 (7) & 0.622 (11) & 0.657 (12) & 0.581 (11) & 0.680 (8) & 0.542 (11) & 0.669 (8) & 0.553 (11) \\
System & Kendall & 0.666 (11) & 0.766 (8) & 0.724 (8) & 0.603 (12) & 0.681 (11) & 0.533 (12) & 0.680 (9) & 0.532 (12) & 0.637 (12) & 0.534 (12) \\
Input & Pearson & 0.901 (1) & 0.876 (3) & 0.887 (1) & 0.843 (1) & 0.854 (7) & 0.844 (2) & 0.842 (5) & 0.736 (7) & 0.828 (2) & 0.784 (5) \\
Input & Spearman & 0.867 (2) & 0.842 (5) & 0.824 (3) & 0.809 (2) & 0.852 (9) & 0.815 (6) & 0.841 (6) & 0.789 (4) & 0.798 (6) & 0.852 (1) \\
Input & Kendall & 0.862 (3) & 0.873 (4) & 0.815 (4) & 0.799 (3) & 0.897 (3) & 0.836 (4) & 0.833 (7) & 0.791 (3) & 0.792 (7) & 0.833 (2) \\
Global & Pearson & 0.843 (5) & 0.889 (1) & 0.876 (2) & 0.791 (4) & 0.880 (4) & 0.833 (5) & 0.854 (4) & 0.752 (6) & 0.820 (4) & 0.825 (3) \\
Global & Spearman & 0.836 (6) & 0.765 (9) & 0.804 (5) & 0.752 (5) & 0.925 (1) & 0.839 (3) & 0.877 (2) & 0.819 (2) & 0.810 (5) & 0.805 (4) \\
Global & Kendall & 0.803 (7) & 0.825 (6) & 0.789 (6) & 0.751 (6) & 0.845 (10) & 0.862 (1) & 0.860 (3) & 0.780 (5) & 0.823 (3) & 0.773 (6) \\
Item & Pearson & 0.679 (9) & 0.775 (7) & 0.679 (10) & 0.746 (7) & 0.861 (6) & 0.749 (9) & 0.669 (10) & 0.652 (9) & 0.657 (9) & 0.741 (7) \\
Item & Spearman & 0.673 (10) & 0.704 (11) & 0.677 (11) & 0.730 (8) & 0.911 (2) & 0.754 (8) & 0.653 (11) & 0.669 (8) & 0.654 (10) & 0.731 (8) \\
Item & Kendall & 0.655 (12) & 0.673 (12) & 0.680 (9) & 0.722 (9) & 0.878 (5) & 0.740 (10) & 0.601 (12) & 0.639 (10) & 0.654 (11) & 0.709 (9) \\ \hline
Grouping & Correlation & D11 & D12 & D13 & D14 & D15 & D16 & D17 & D18 & D19 & D20 \\ \hline
System & Pearson & 0.822 (1) & 0.595 (12) & 0.554 (10) & 0.674 (7) & 0.557 (12) & 0.671 (6) & 0.589 (12) & 0.000 (12) & 0.613 (2) & 0.782 (5) \\
System & Spearman & 0.584 (10) & 0.771 (7) & 0.473 (12) & 0.889 (1) & 0.562 (10) & 0.538 (12) & 0.751 (6) & 0.259 (10) & 0.606 (3) & 0.132 (12) \\
System & Kendall & 0.569 (12) & 0.771 (8) & 0.487 (11) & 0.886 (2) & 0.558 (11) & 0.592 (11) & 0.752 (5) & 0.150 (11) & 0.619 (1) & 0.148 (11) \\
Input & Pearson & 0.755 (5) & 0.803 (5) & 0.695 (4) & 0.684 (6) & 0.777 (3) & 0.689 (4) & 0.809 (3) & 0.382 (7) & 0.336 (4) & 0.853 (1) \\
Input & Spearman & 0.795 (4) & 0.788 (6) & 0.685 (5) & 0.670 (8) & 0.748 (6) & 0.679 (5) & 0.737 (7) & 0.284 (9) & 0.196 (5) & 0.771 (6) \\
Input & Kendall & 0.733 (6) & 0.805 (4) & 0.671 (6) & 0.651 (9) & 0.723 (8) & 0.659 (7) & 0.723 (9) & 0.318 (8) & 0.192 (6) & 0.763 (7) \\
Global & Pearson & 0.720 (7) & 0.859 (2) & 0.770 (3) & 0.807 (3) & 0.745 (7) & 0.784 (1) & 0.826 (2) & 0.662 (1) & -0.082 (9) & 0.841 (2) \\
Global & Spearman & 0.803 (2) & 0.878 (1) & 0.772 (2) & 0.745 (5) & 0.828 (1) & 0.780 (2) & 0.834 (1) & 0.588 (3) & 0.065 (7) & 0.823 (3) \\
Global & Kendall & 0.798 (3) & 0.834 (3) & 0.782 (1) & 0.747 (4) & 0.792 (2) & 0.777 (3) & 0.759 (4) & 0.630 (2) & 0.053 (8) & 0.816 (4) \\
Item & Pearson & 0.606 (8) & 0.701 (10) & 0.574 (9) & 0.580 (11) & 0.765 (4) & 0.644 (9) & 0.724 (8) & 0.560 (4) & -0.112 (10) & 0.618 (8) \\
Item & Spearman & 0.590 (9) & 0.712 (9) & 0.604 (7) & 0.583 (10) & 0.750 (5) & 0.659 (8) & 0.716 (10) & 0.541 (6) & -0.150 (11) & 0.570 (9) \\
Item & Kendall & 0.573 (11) & 0.648 (11) & 0.574 (8) & 0.562 (12) & 0.721 (9) & 0.623 (10) & 0.663 (11) & 0.550 (5) & -0.155 (12) & 0.552 (10) \\ \hline
Grouping & Correlation & D21 & D22 & D23 & D24 & D25 & D26 & D27 & D28 & D29 & D30 \\ \hline
System & Pearson & 0.721 (7) & 0.748 (6) & 0.721 (7) & 0.725 (7) & 0.655 (1) & 0.761 (7) & 0.847 (2) & 0.596 (10) & 0.725 (7) & 0.663 (10) \\
System & Spearman & 0.179 (11) & 0.169 (12) & 0.085 (12) & 0.115 (11) & 0.177 (11) & 0.295 (12) & 0.698 (11) & 0.373 (12) & 0.039 (12) & 0.390 (12) \\
System & Kendall & 0.150 (12) & 0.184 (11) & 0.092 (11) & 0.113 (12) & 0.167 (12) & 0.391 (11) & 0.722 (10) & 0.391 (11) & 0.062 (11) & 0.418 (11) \\
Input & Pearson & 0.794 (3) & 0.809 (2) & 0.823 (2) & 0.827 (1) & 0.518 (5) & 0.803 (3) & 0.802 (4) & 0.861 (2) & 0.786 (2) & 0.846 (2) \\
Input & Spearman & 0.728 (6) & 0.781 (4) & 0.813 (4) & 0.754 (6) & 0.503 (7) & 0.778 (5) & 0.765 (6) & 0.849 (4) & 0.726 (6) & 0.807 (6) \\
Input & Kendall & 0.704 (9) & 0.775 (5) & 0.781 (6) & 0.758 (5) & 0.514 (6) & 0.803 (2) & 0.806 (3) & 0.882 (1) & 0.749 (3) & 0.843 (3) \\
Global & Pearson & 0.774 (4) & 0.804 (3) & 0.834 (1) & 0.774 (4) & 0.523 (4) & 0.835 (1) & 0.857 (1) & 0.854 (3) & 0.835 (1) & 0.842 (4) \\
Global & Spearman & 0.809 (2) & 0.810 (1) & 0.816 (3) & 0.789 (2) & 0.530 (3) & 0.799 (4) & 0.775 (5) & 0.830 (5) & 0.738 (5) & 0.855 (1) \\
Global & Kendall & 0.829 (1) & 0.712 (7) & 0.809 (5) & 0.780 (3) & 0.542 (2) & 0.763 (6) & 0.740 (7) & 0.784 (6) & 0.709 (8) & 0.818 (5) \\
Item & Pearson & 0.742 (5) & 0.665 (8) & 0.658 (8) & 0.657 (9) & 0.284 (8) & 0.686 (10) & 0.739 (8) & 0.783 (7) & 0.746 (4) & 0.761 (8) \\
Item & Spearman & 0.699 (10) & 0.623 (9) & 0.658 (9) & 0.674 (8) & 0.202 (9) & 0.751 (8) & 0.735 (9) & 0.769 (8) & 0.668 (9) & 0.782 (7) \\
Item & Kendall & 0.717 (8) & 0.541 (10) & 0.656 (10) & 0.656 (10) & 0.185 (10) & 0.698 (9) & 0.676 (12) & 0.736 (9) & 0.663 (10) & 0.751 (9) \\ \hline
\end{tabular}
}
\caption{The complete RC values of different correlation measures on all meta-evaluation datasets, the higher the better, which are visualized as Figure \ref{fig:RC}.} 
\label{tab:RC}
\end{table*}

\clearpage

\begin{figure}[p]
  \includegraphics[width=\linewidth]{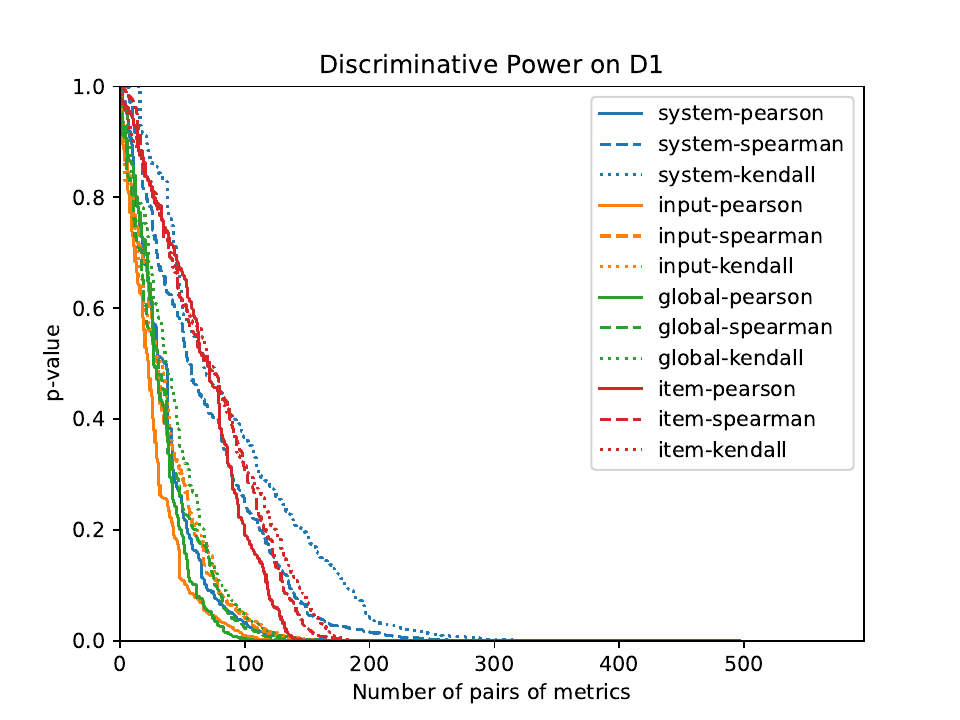} \hfill
  \caption {The p-value curves of correlation measures on meta-evaluation D1.}
  \label{fig:p_value_curve_D1}
\end{figure}

\begin{figure}[p]
  \includegraphics[width=\linewidth]{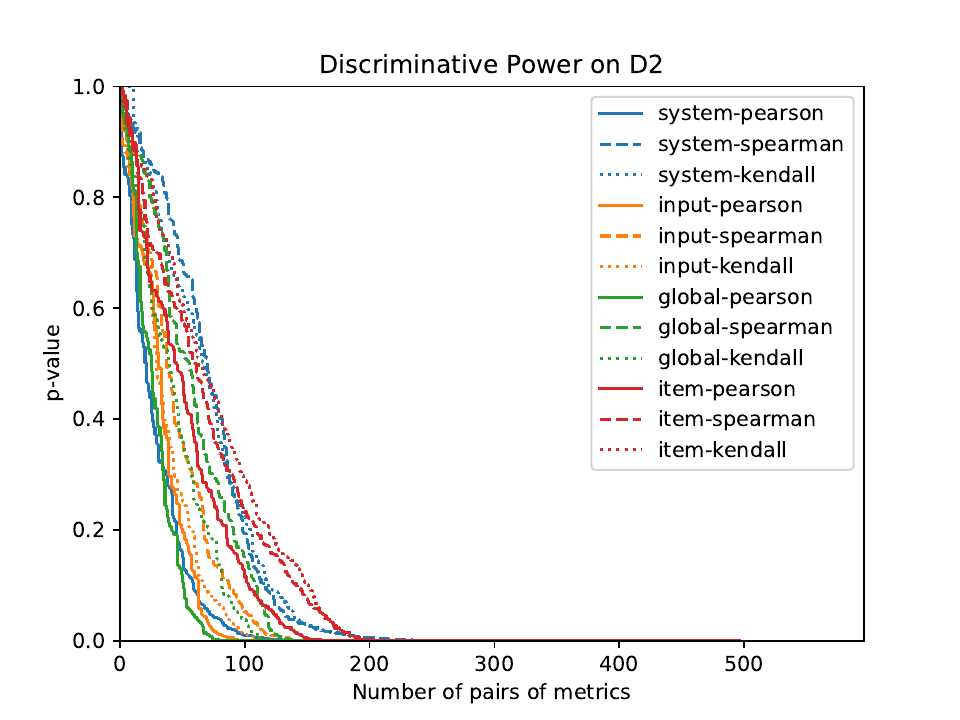} \hfill
  \caption {The p-value curves of correlation measures on meta-evaluation D2.}
  \label{fig:p_value_curve_D2}
\end{figure}

\begin{figure}[p]
  \includegraphics[width=\linewidth]{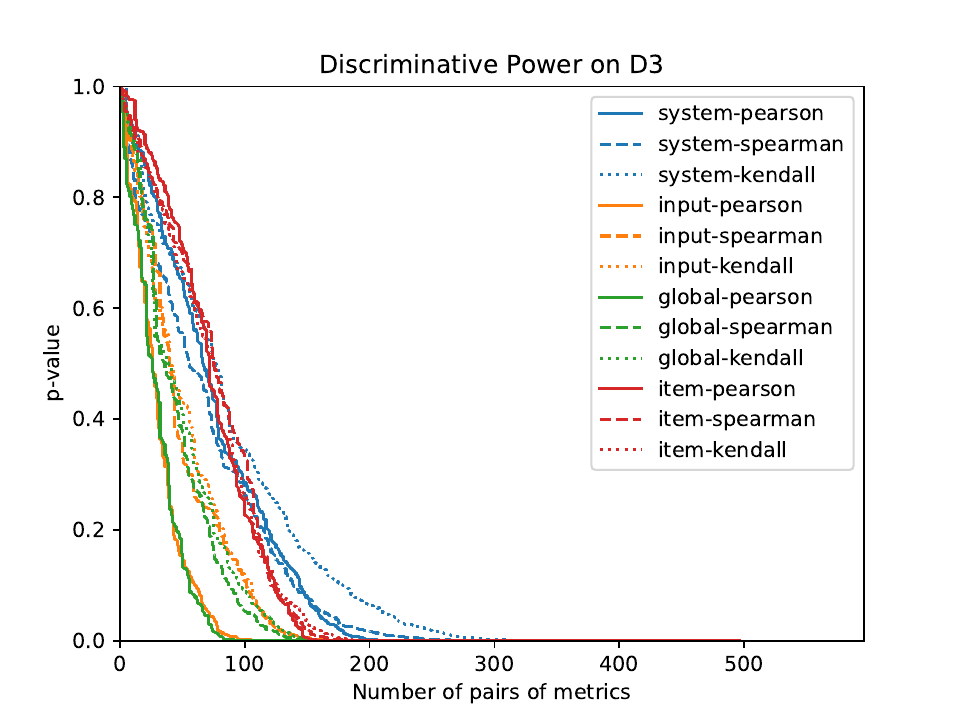} \hfill
  \caption {The p-value curves of correlation measures on meta-evaluation D3.}
  \label{fig:p_value_curve_D3}
\end{figure}

\begin{figure}[p]
  \includegraphics[width=\linewidth]{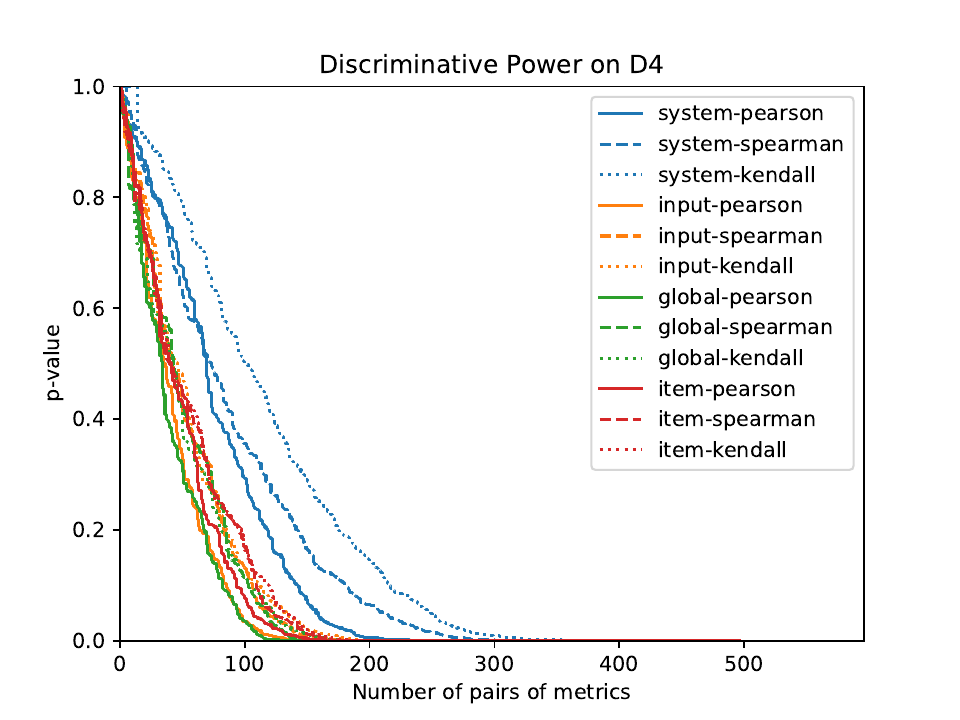} \hfill
  \caption {The p-value curves of correlation measures on meta-evaluation D4.}
  \label{fig:p_value_curve_D4}
\end{figure}

\begin{figure}[p]
  \includegraphics[width=\linewidth]{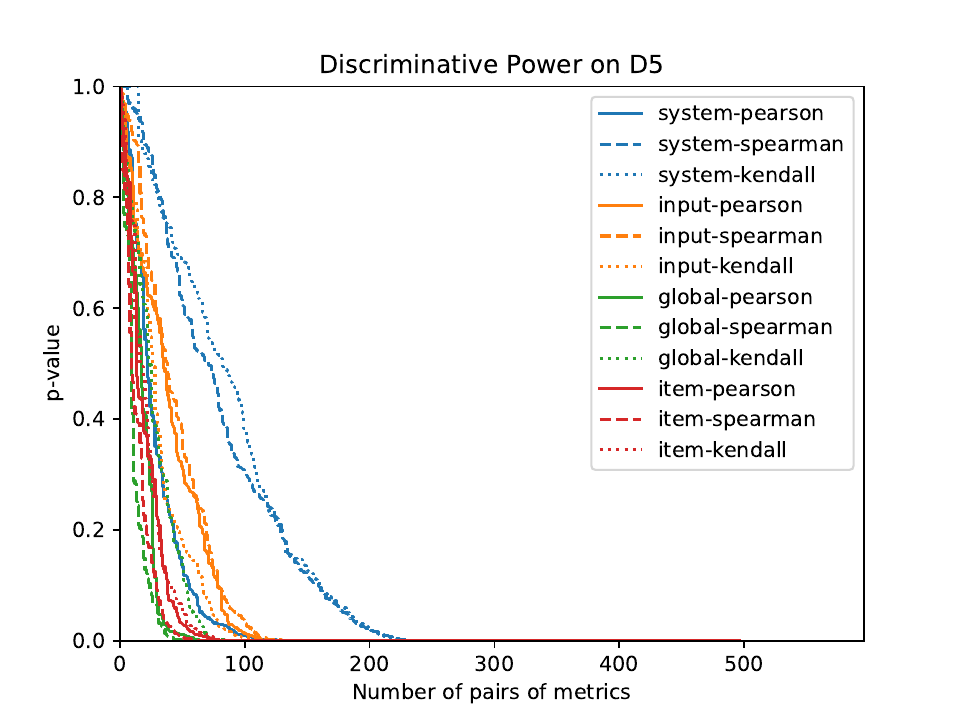} \hfill
  \caption {The p-value curves of correlation measures on meta-evaluation D5.}
  \label{fig:p_value_curve_D5}
\end{figure}

\begin{figure}[p]
  \includegraphics[width=\linewidth]{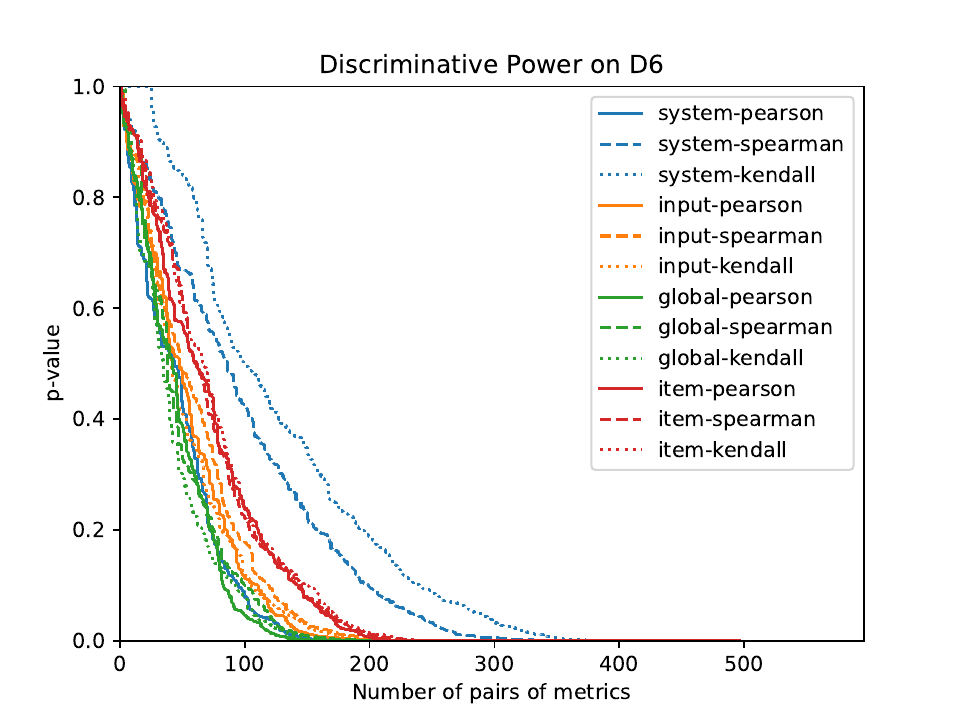} \hfill
  \caption {The p-value curves of correlation measures on meta-evaluation D6.}
  \label{fig:p_value_curve_D6}
\end{figure}

\begin{figure}[!htp]
  \includegraphics[width=\linewidth]{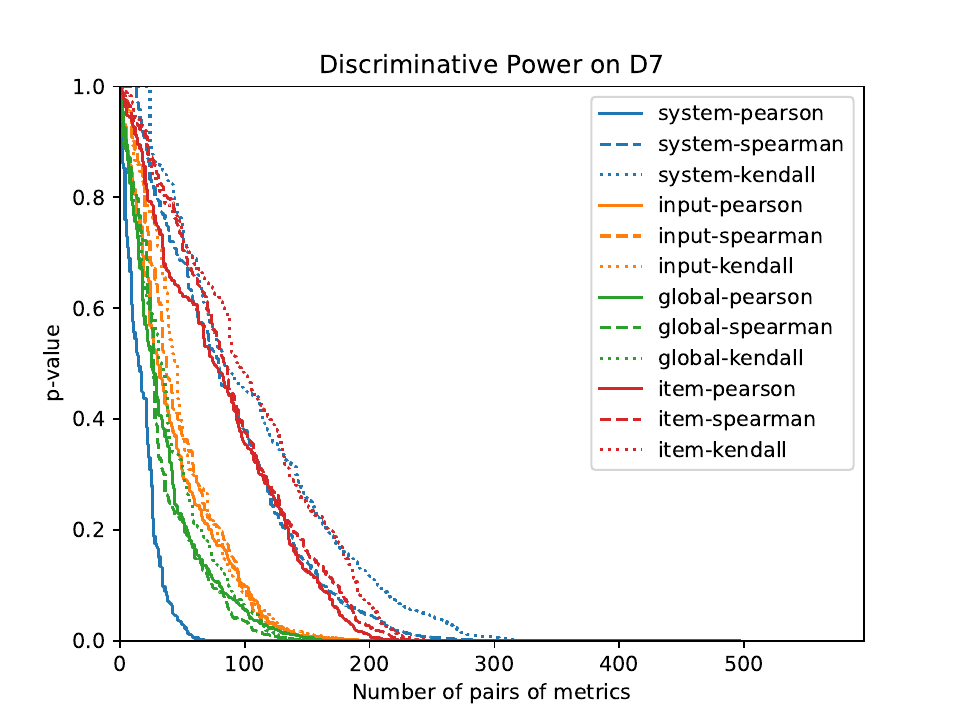} \hfill
  \caption {The p-value curves of correlation measures on meta-evaluation D7.}
  \label{fig:p_value_curve_D7}
\end{figure}

\begin{figure}[!htp]
  \includegraphics[width=\linewidth]{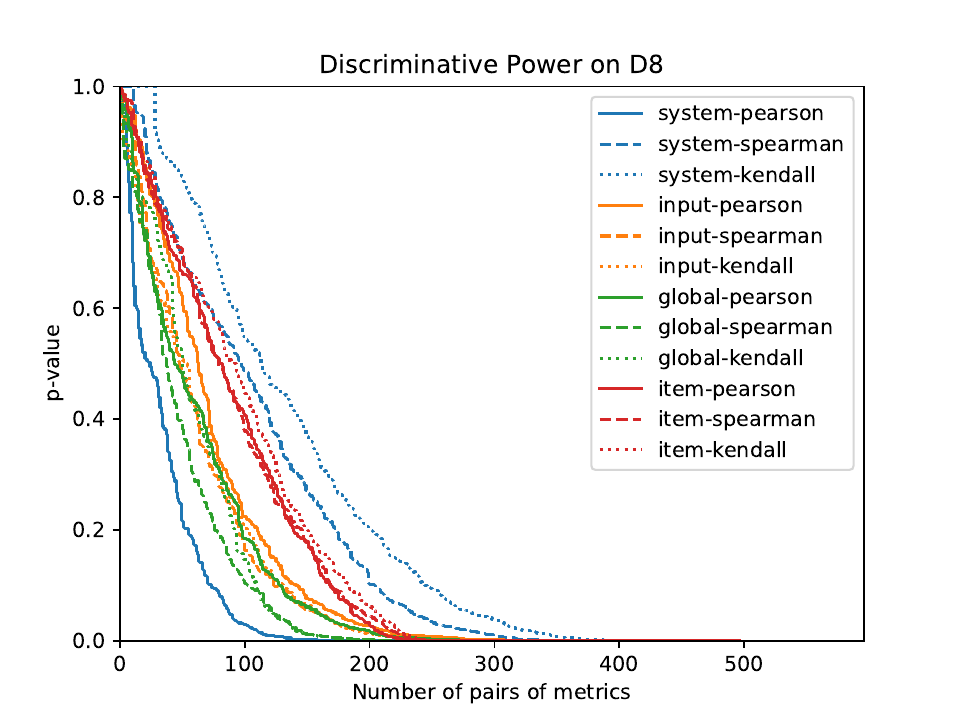} \hfill
  \caption {The p-value curves of correlation measures on meta-evaluation D8.}
  \label{fig:p_value_curve_D8}
\end{figure}

\begin{figure}[!htp]
  \includegraphics[width=\linewidth]{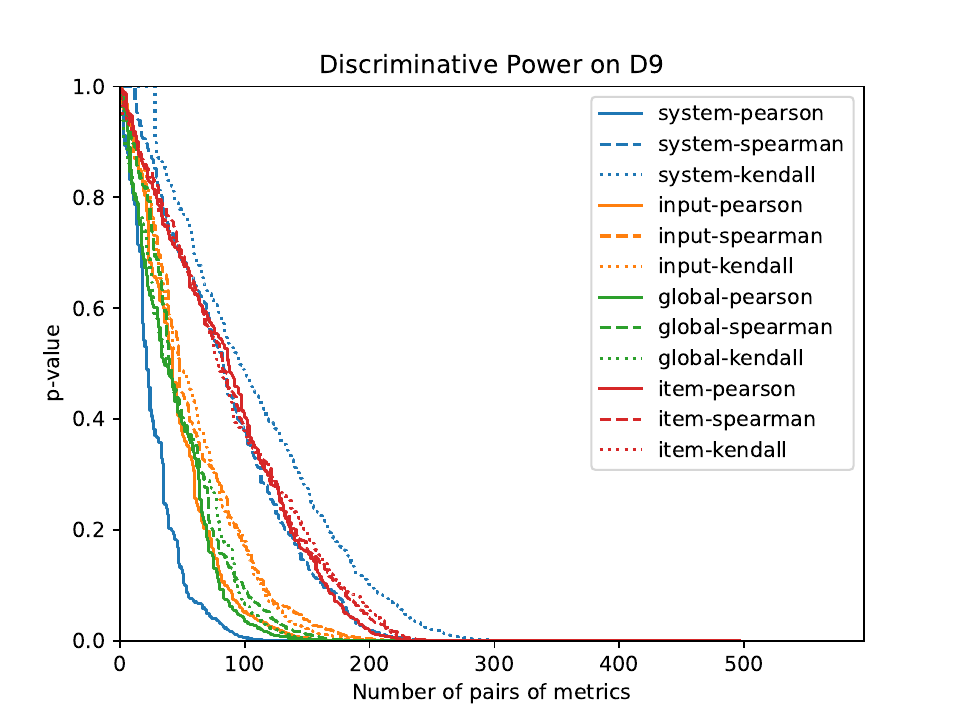} \hfill
  \caption {The p-value curves of correlation measures on meta-evaluation D9.}
  \label{fig:p_value_curve_D9}
\end{figure}

\begin{figure}[!htp]
  \includegraphics[width=\linewidth]{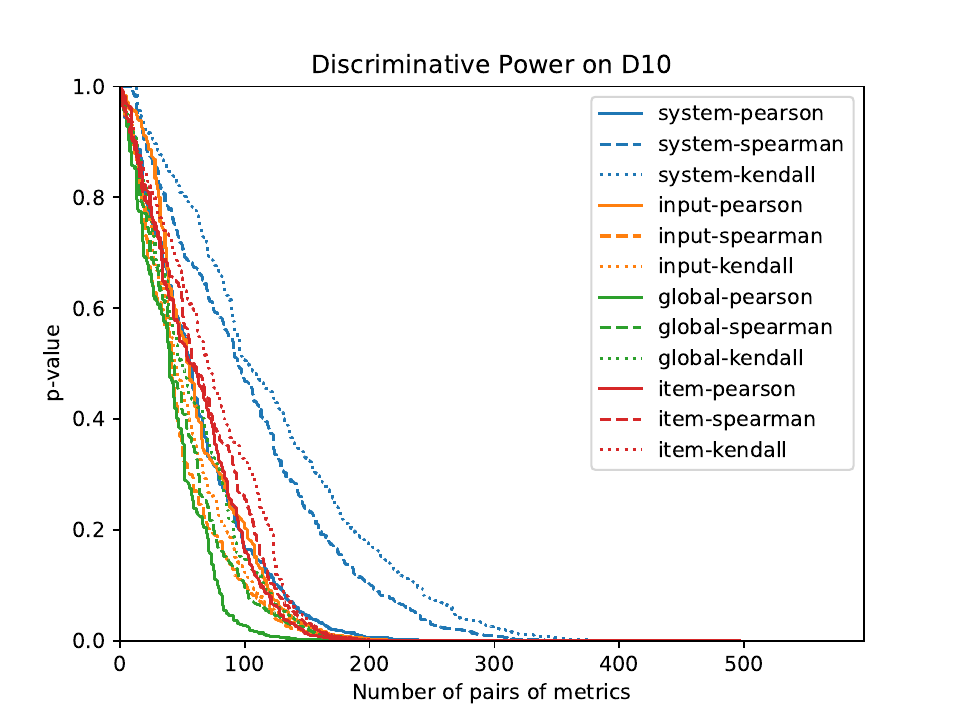} \hfill
  \caption {The p-value curves of correlation measures on meta-evaluation D10.}
  \label{fig:p_value_curve_D10}
\end{figure}

\begin{figure}[!htp]
  \includegraphics[width=\linewidth]{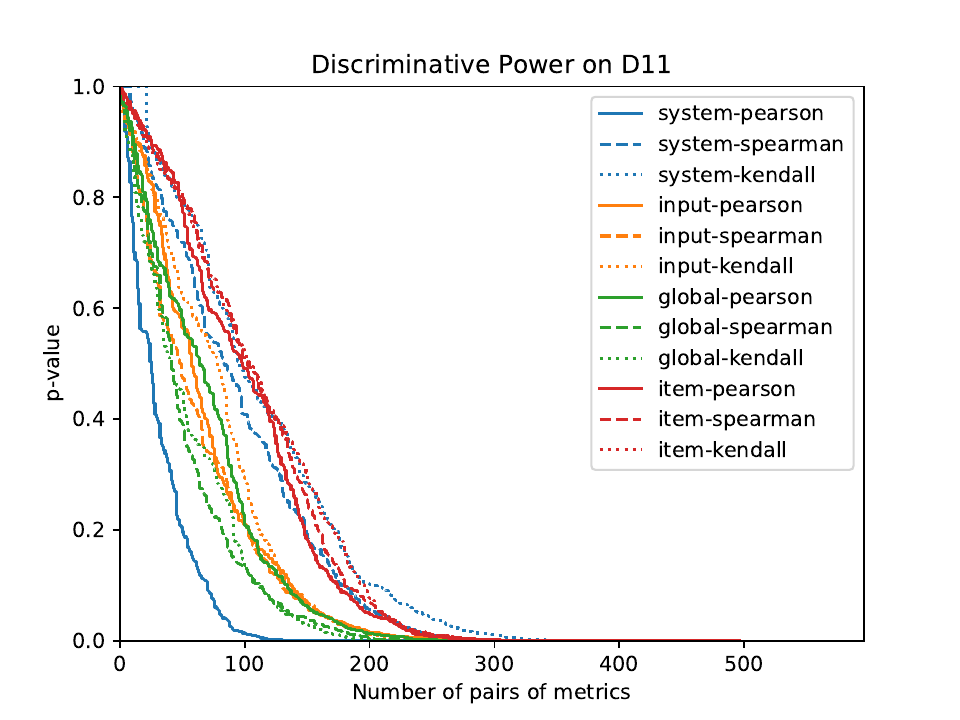} \hfill
  \caption {The p-value curves of correlation measures on meta-evaluation D11.}
  \label{fig:p_value_curve_D11}
\end{figure}

\begin{figure}[!htp]
  \includegraphics[width=\linewidth]{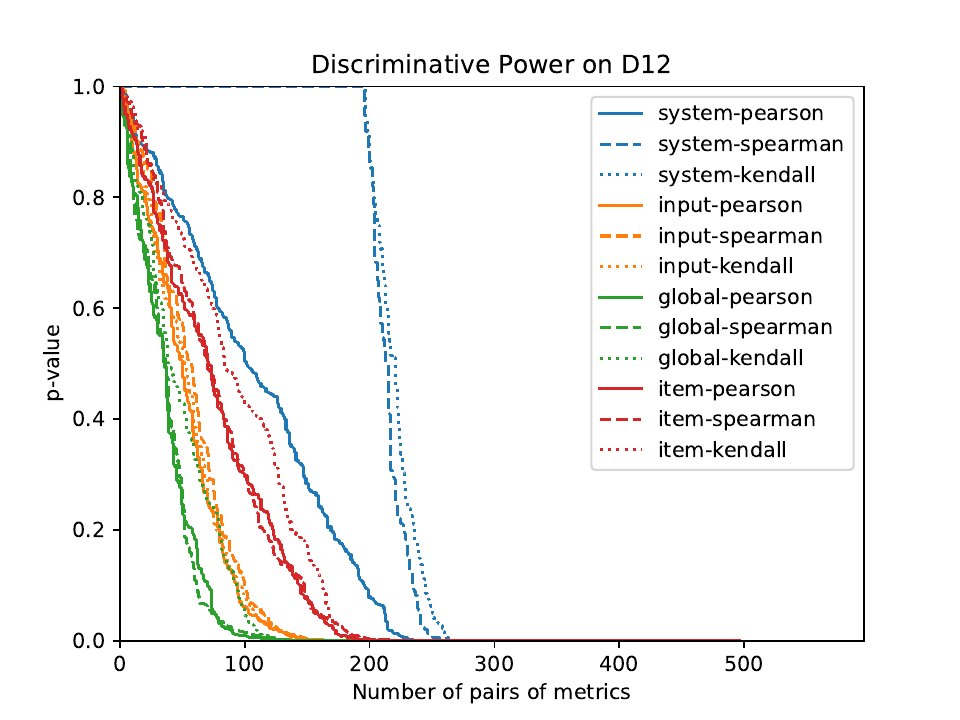} \hfill
  \caption {The p-value curves of correlation measures on meta-evaluation D12.}
  \label{fig:p_value_curve_D12}
\end{figure}

\begin{figure}[!htp]
  \includegraphics[width=\linewidth]{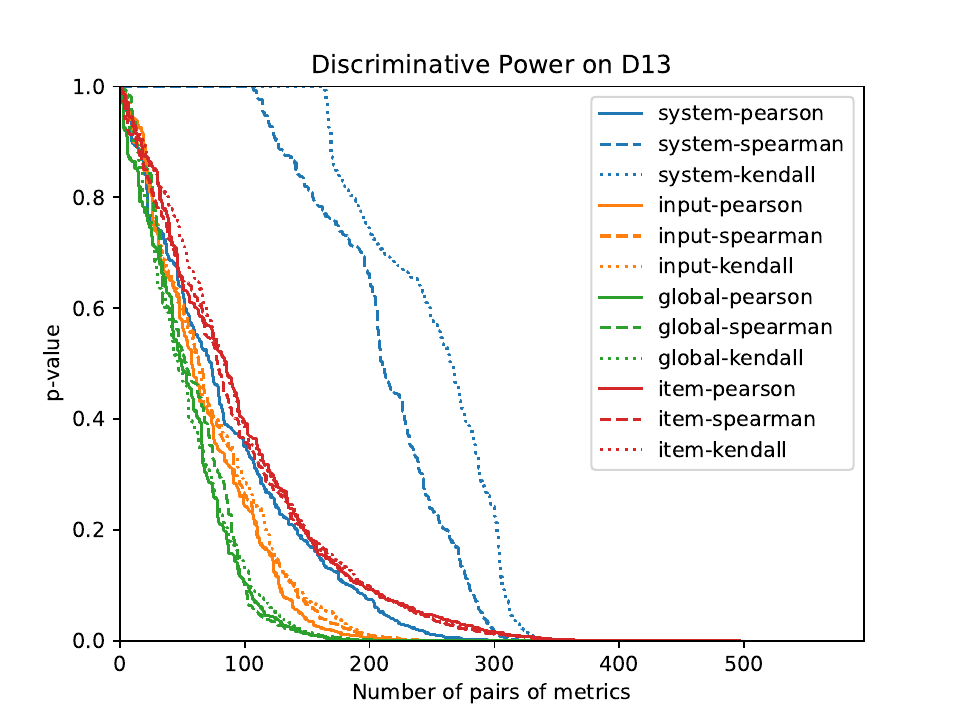} \hfill
  \caption {The p-value curves of correlation measures on meta-evaluation D13.}
  \label{fig:p_value_curve_D13}
\end{figure}

\begin{figure}[!htp]
  \includegraphics[width=\linewidth]{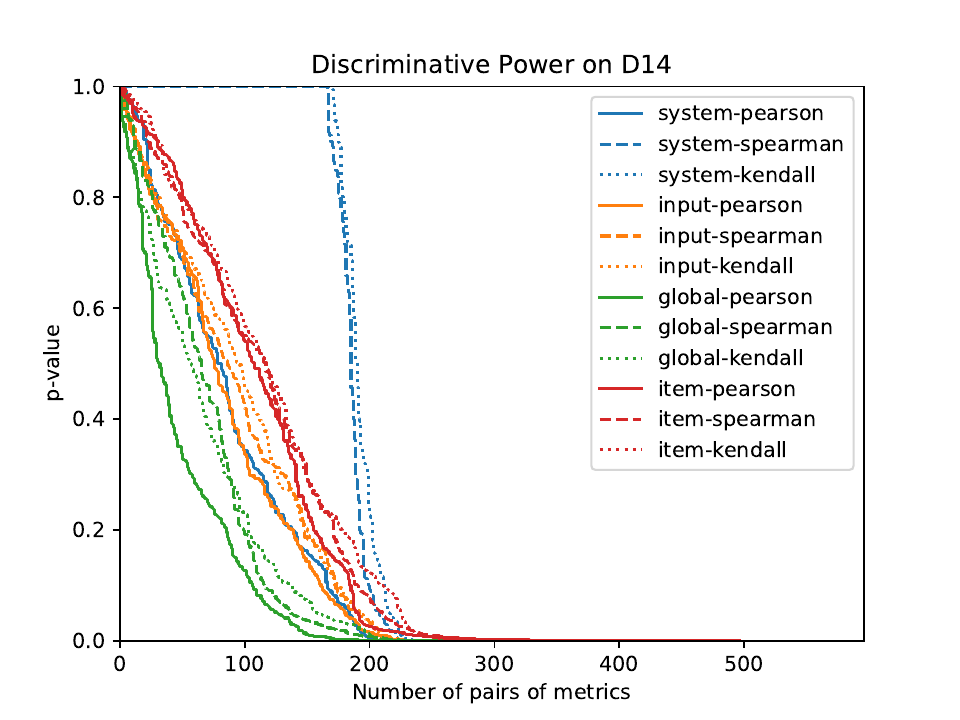} \hfill
  \caption {The p-value curves of correlation measures on meta-evaluation D14.}
  \label{fig:p_value_curve_D14}
\end{figure}

\begin{figure}[!htp]
  \includegraphics[width=\linewidth]{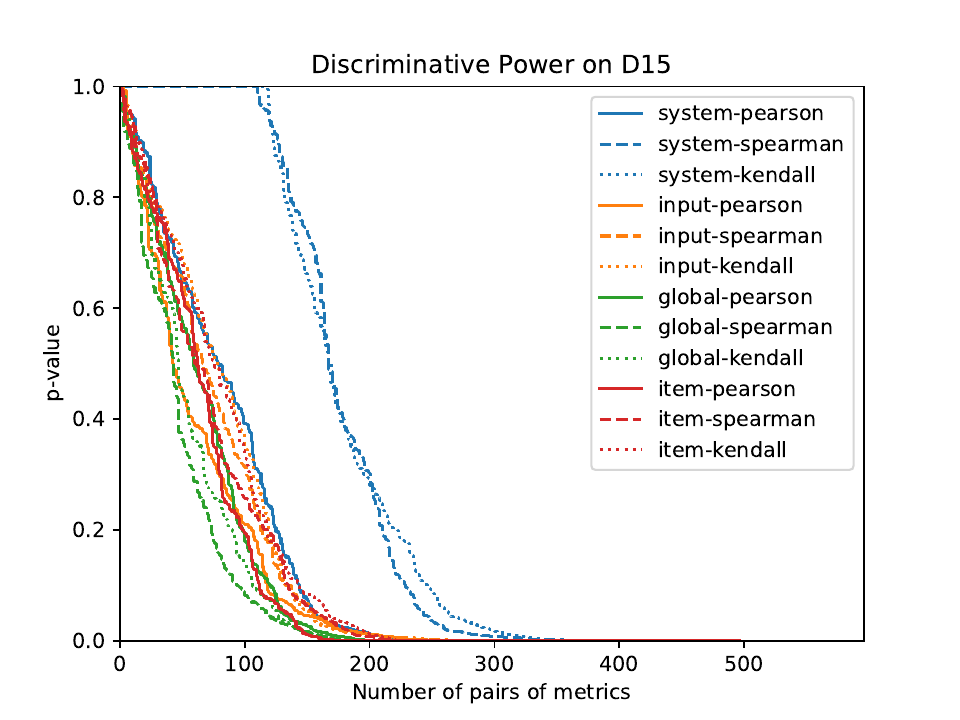} \hfill
  \caption {The p-value curves of correlation measures on meta-evaluation D15.}
  \label{fig:p_value_curve_D15}
\end{figure}

\begin{figure}[!htp]
  \includegraphics[width=\linewidth]{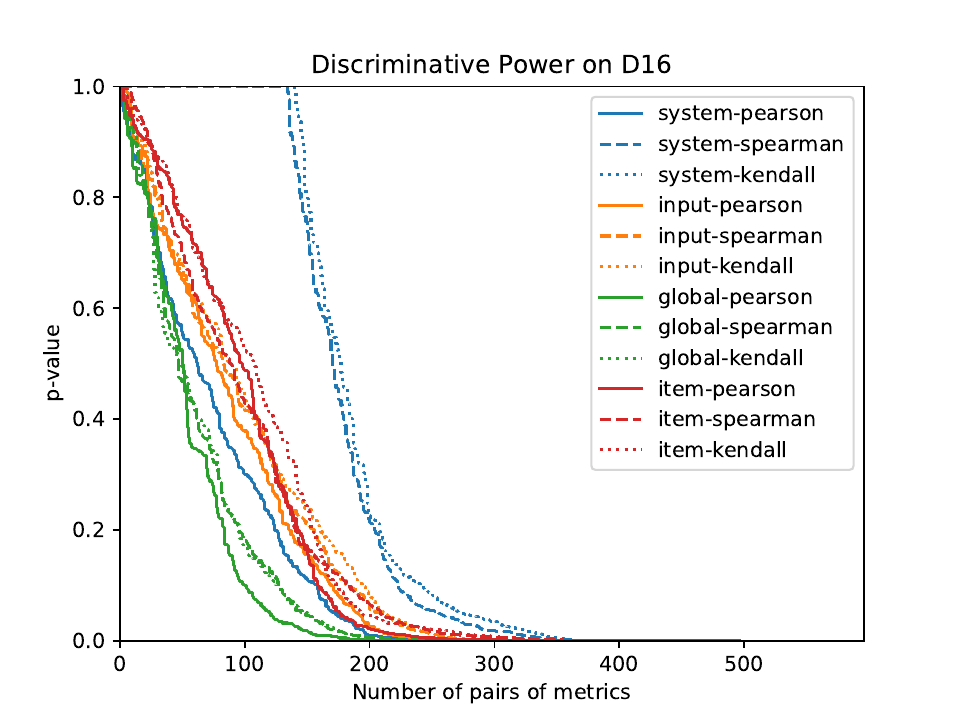} \hfill
  \caption {The p-value curves of correlation measures on meta-evaluation D16.}
  \label{fig:p_value_curve_D16}
\end{figure}

\begin{figure}[!htp]
  \includegraphics[width=\linewidth]{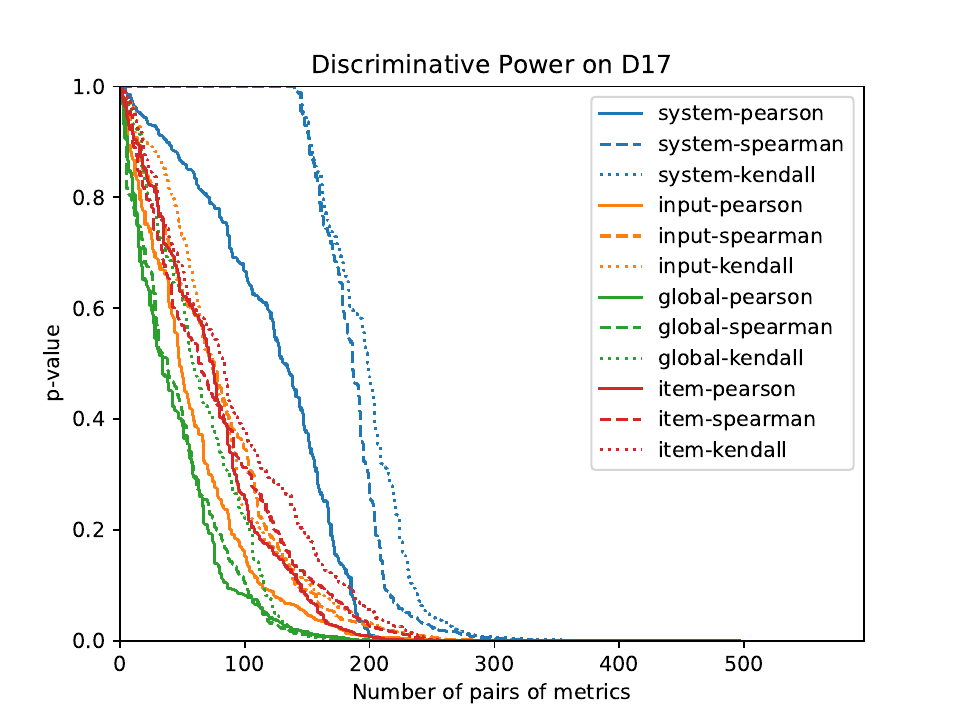} \hfill
  \caption {The p-value curves of correlation measures on meta-evaluation D17.}
  \label{fig:p_value_curve_D17}
\end{figure}

\begin{figure}[!htp]
  \includegraphics[width=\linewidth]{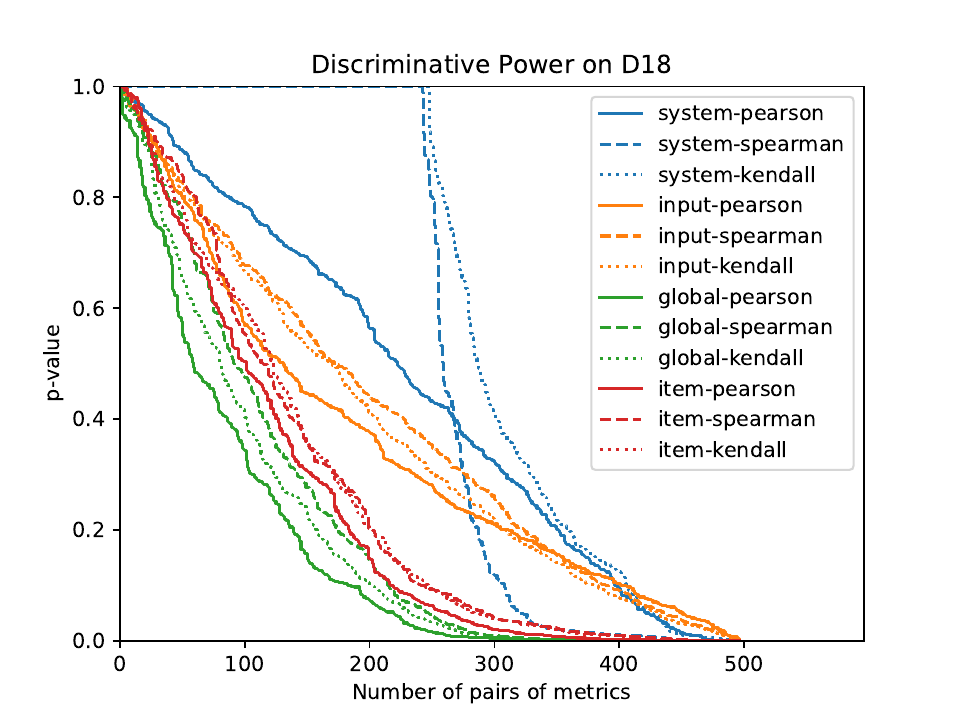} \hfill
  \caption {The p-value curves of correlation measures on meta-evaluation D18.}
  \label{fig:p_value_curve_D18}
\end{figure}

\begin{figure}[!htp]
  \includegraphics[width=\linewidth]{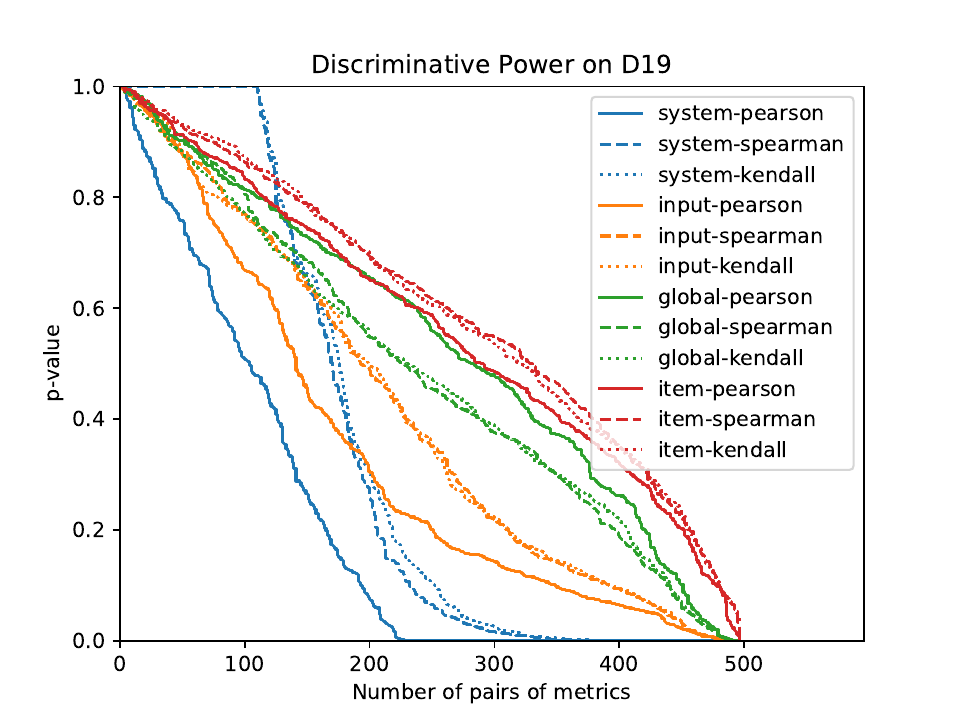} \hfill
  \caption {The p-value curves of correlation measures on meta-evaluation D19.}
  \label{fig:p_value_curve_D19}
\end{figure}

\begin{figure}[!htp]
  \includegraphics[width=\linewidth]{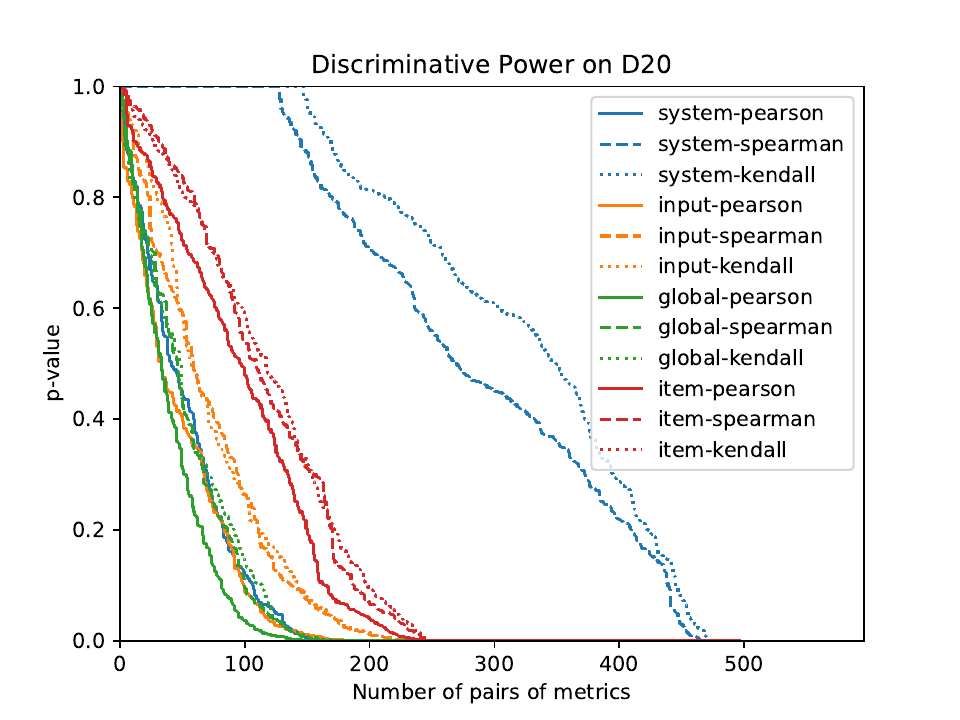} \hfill
  \caption {The p-value curves of correlation measures on meta-evaluation D20.}
  \label{fig:p_value_curve_D20}
\end{figure}

\begin{figure}[!htp]
  \includegraphics[width=\linewidth]{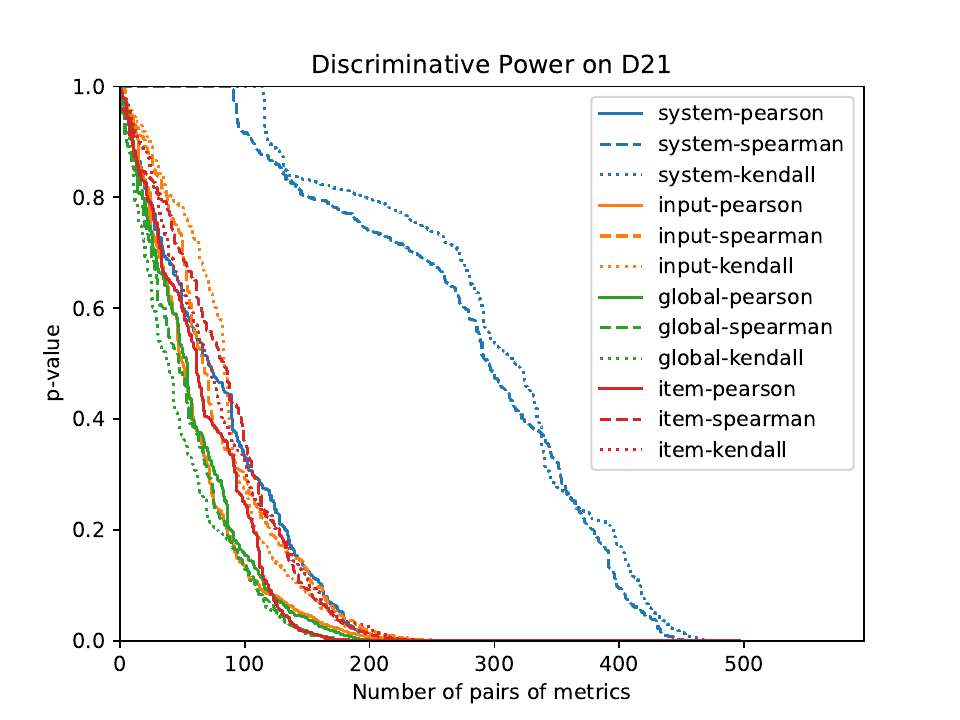} \hfill
  \caption {The p-value curves of correlation measures on meta-evaluation D21.}
  \label{fig:p_value_curve_D21}
\end{figure}

\begin{figure}[!htp]
  \includegraphics[width=\linewidth]{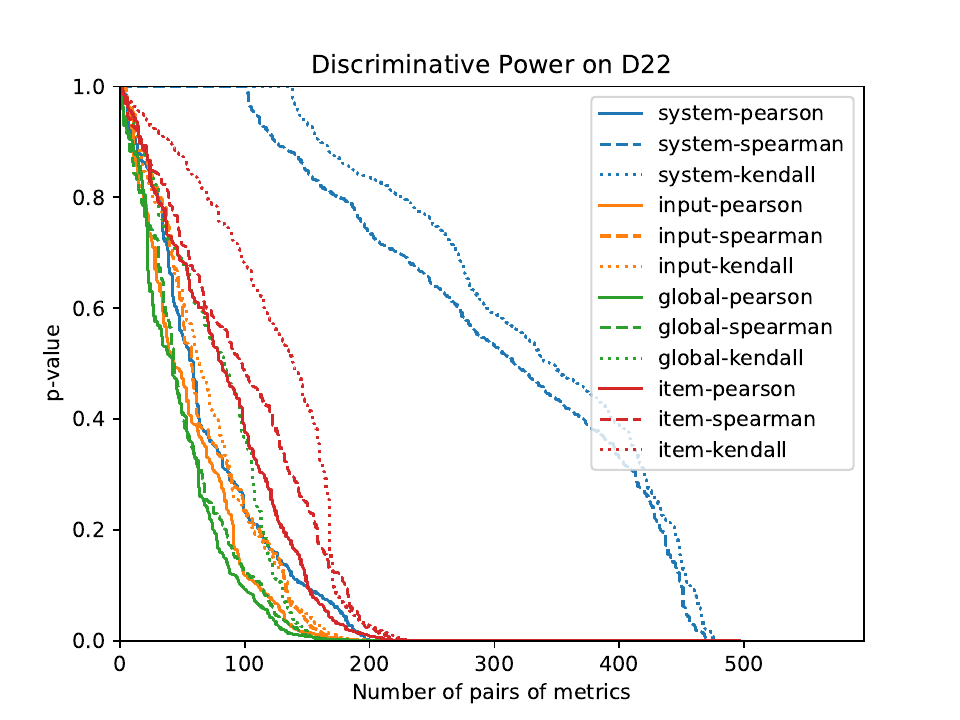} \hfill
  \caption {The p-value curves of correlation measures on meta-evaluation D22.}
  \label{fig:p_value_curve_D22}
\end{figure}

\begin{figure}[!htp]
  \includegraphics[width=\linewidth]{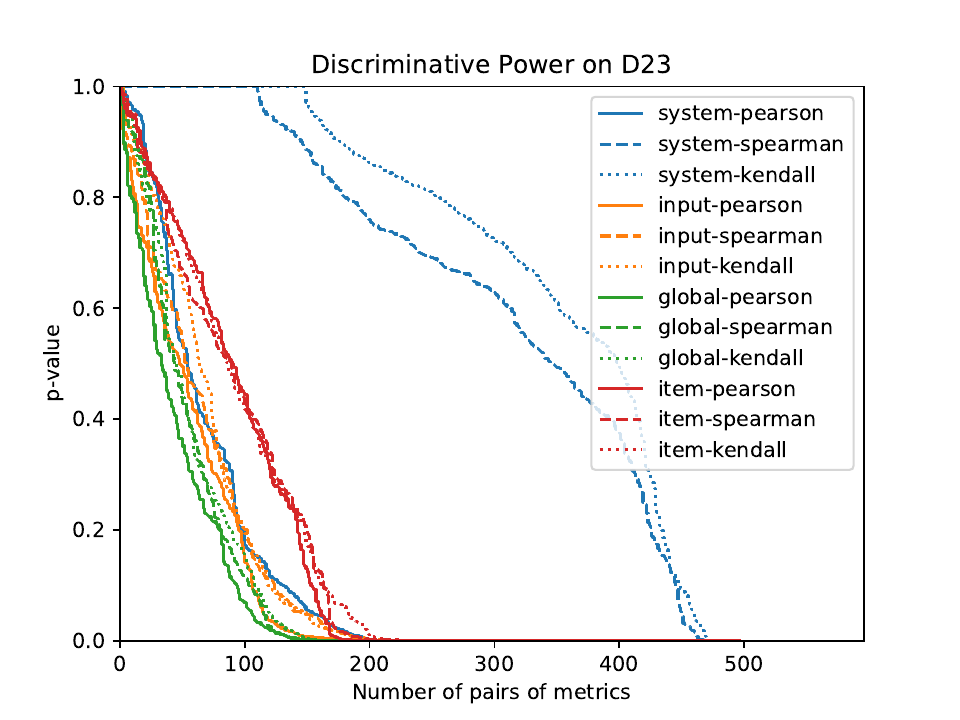} \hfill
  \caption {The p-value curves of correlation measures on meta-evaluation D23.}
  \label{fig:p_value_curve_D23}
\end{figure}

\begin{figure}[!htp]
  \includegraphics[width=\linewidth]{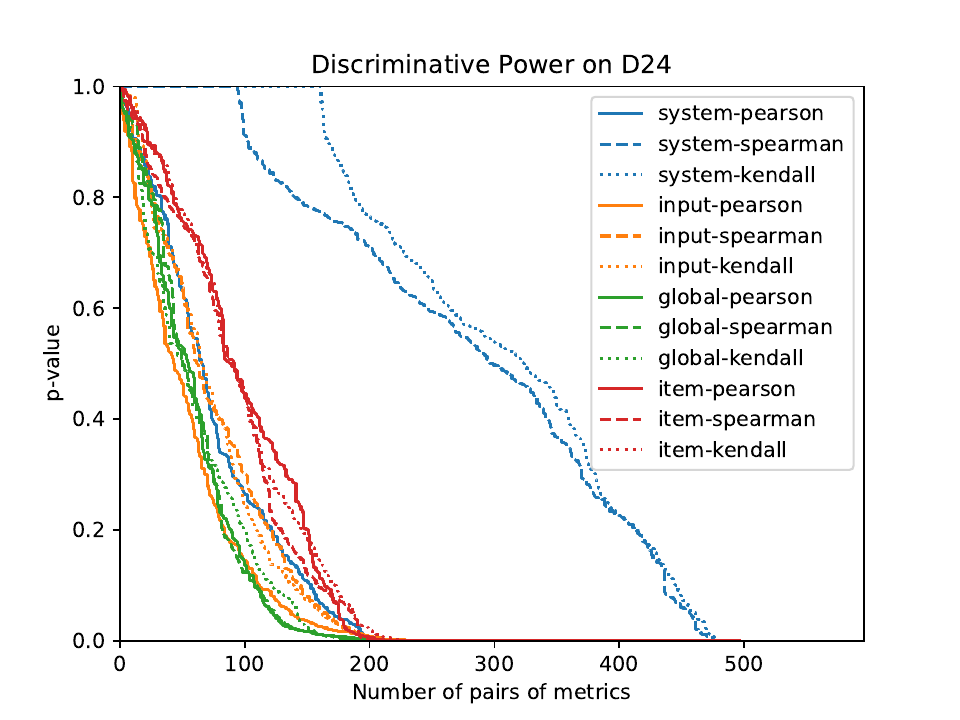} \hfill
  \caption {The p-value curves of correlation measures on meta-evaluation D24.}
  \label{fig:p_value_curve_D24}
\end{figure}

\begin{figure}[!htp]
  \includegraphics[width=\linewidth]{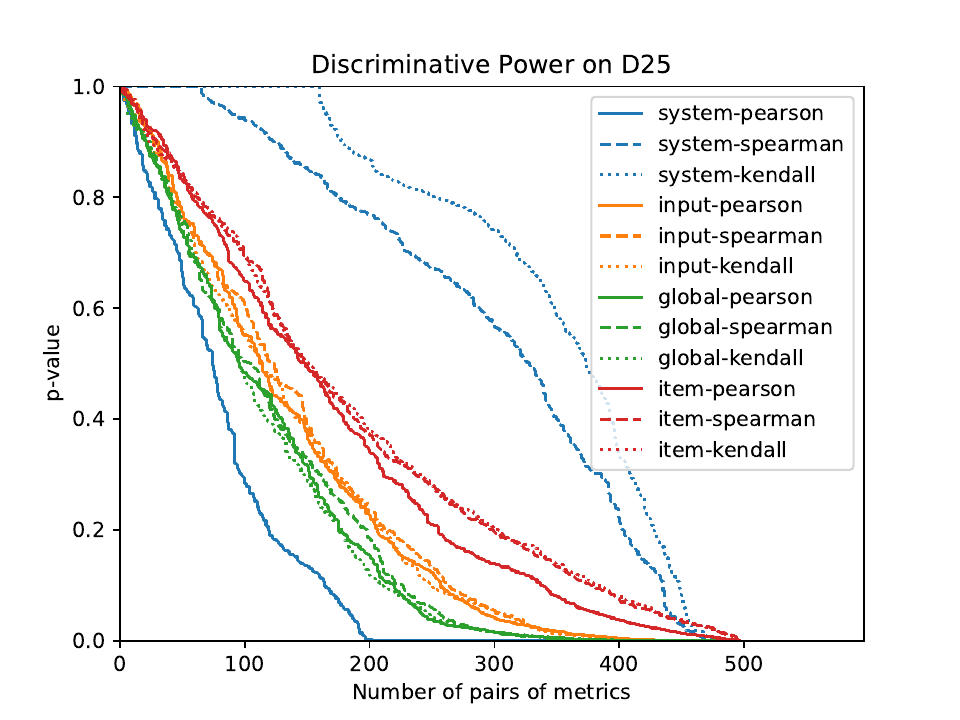} \hfill
  \caption {The p-value curves of correlation measures on meta-evaluation D25.}
  \label{fig:p_value_curve_D25}
\end{figure}

\begin{figure}[!htp]
  \includegraphics[width=\linewidth]{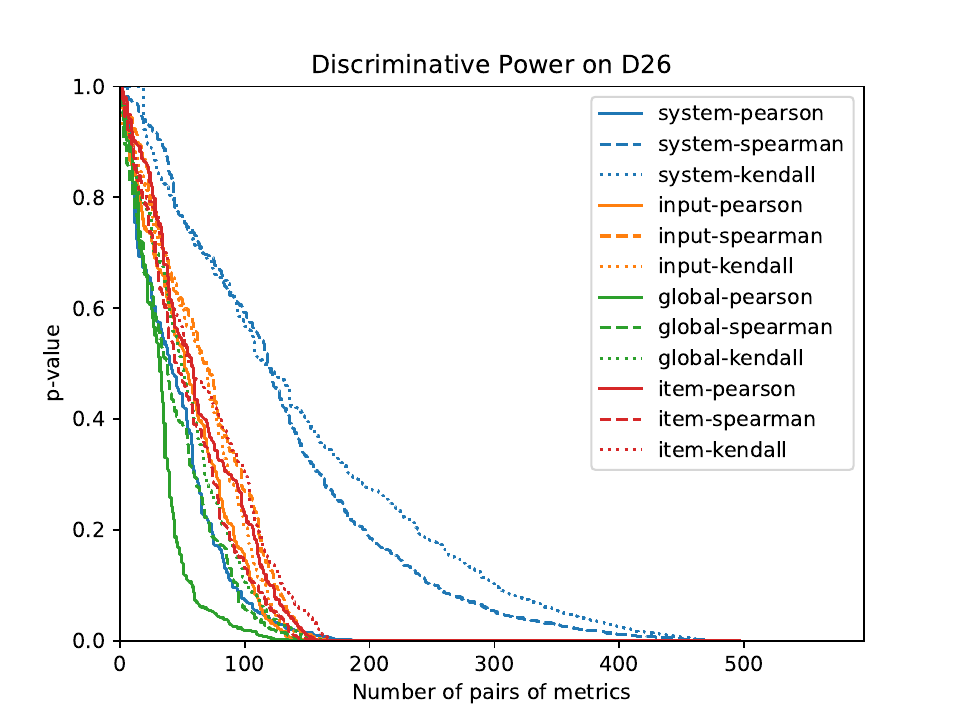} \hfill
  \caption {The p-value curves of correlation measures on meta-evaluation D26.}
  \label{fig:p_value_curve_D26}
\end{figure}

\begin{figure}[!htp]
  \includegraphics[width=\linewidth]{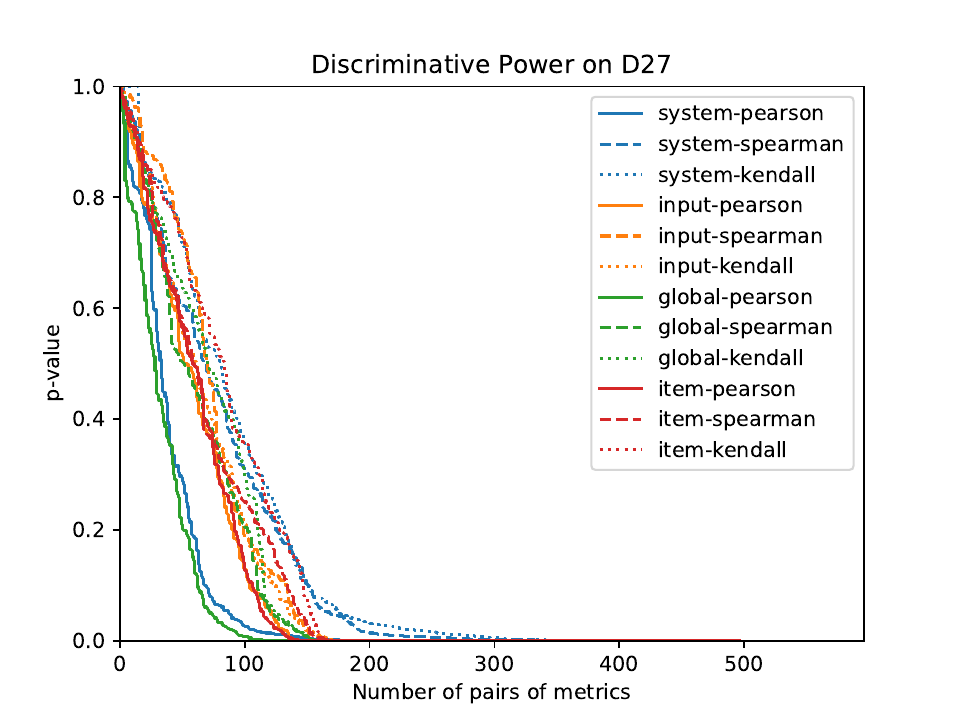} \hfill
  \caption {The p-value curves of correlation measures on meta-evaluation D27.}
  \label{fig:p_value_curve_D27}
\end{figure}

\begin{figure}[!htp]
  \includegraphics[width=\linewidth]{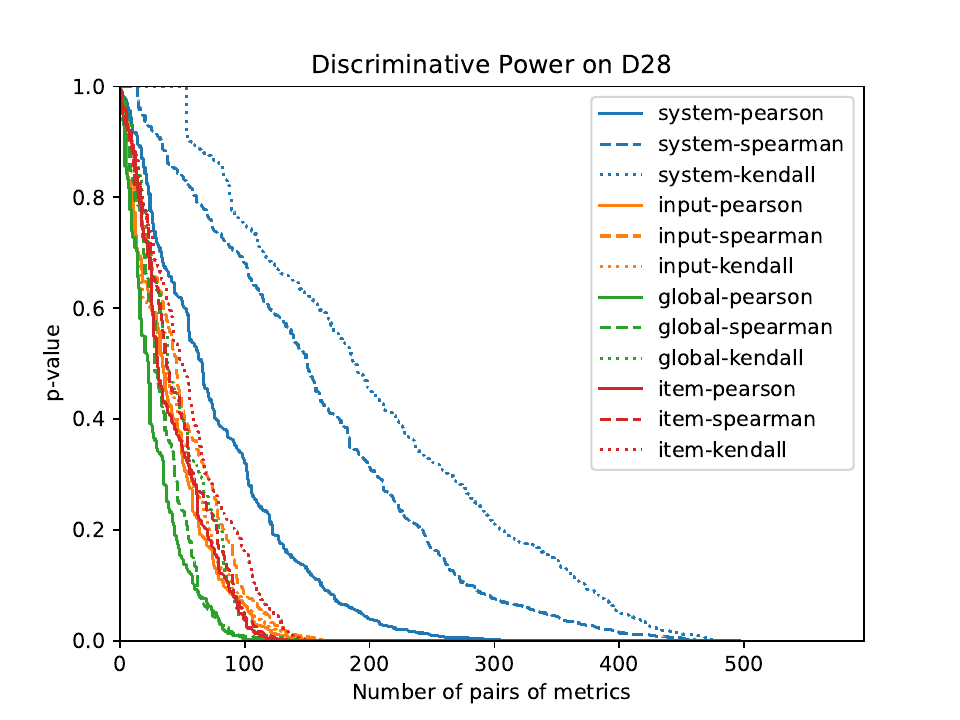} \hfill
  \caption {The p-value curves of correlation measures on meta-evaluation D28.}
  \label{fig:p_value_curve_D28}
\end{figure}

\begin{figure}[!htp]
  \includegraphics[width=\linewidth]{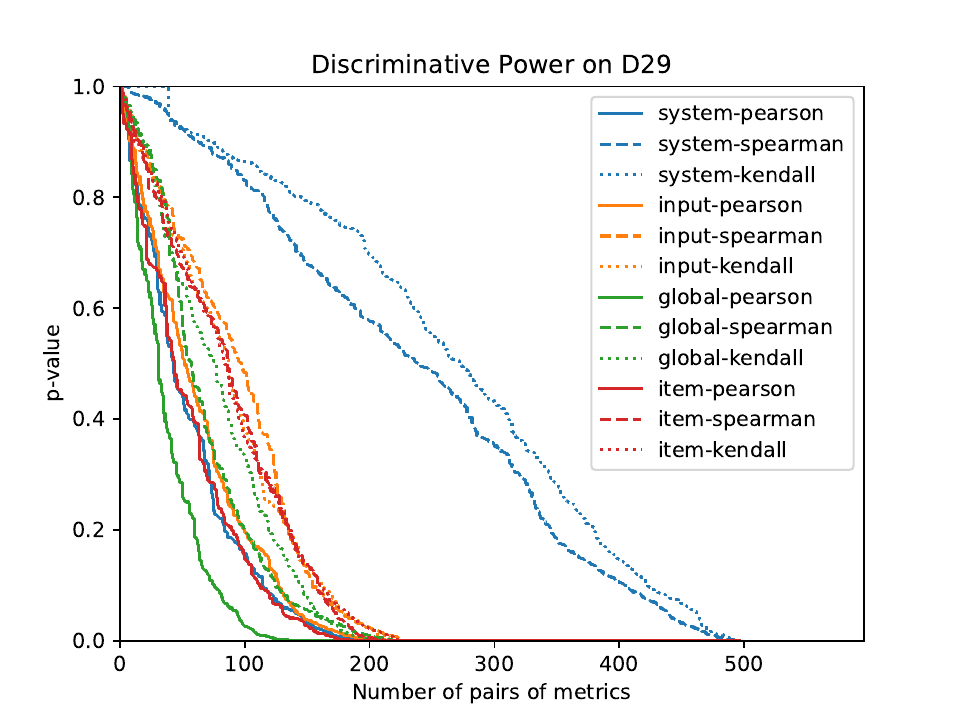} \hfill
  \caption {The p-value curves of correlation measures on meta-evaluation D29.}
  \label{fig:p_value_curve_D29}
\end{figure}

\begin{figure}[!htp]
  \includegraphics[width=\linewidth]{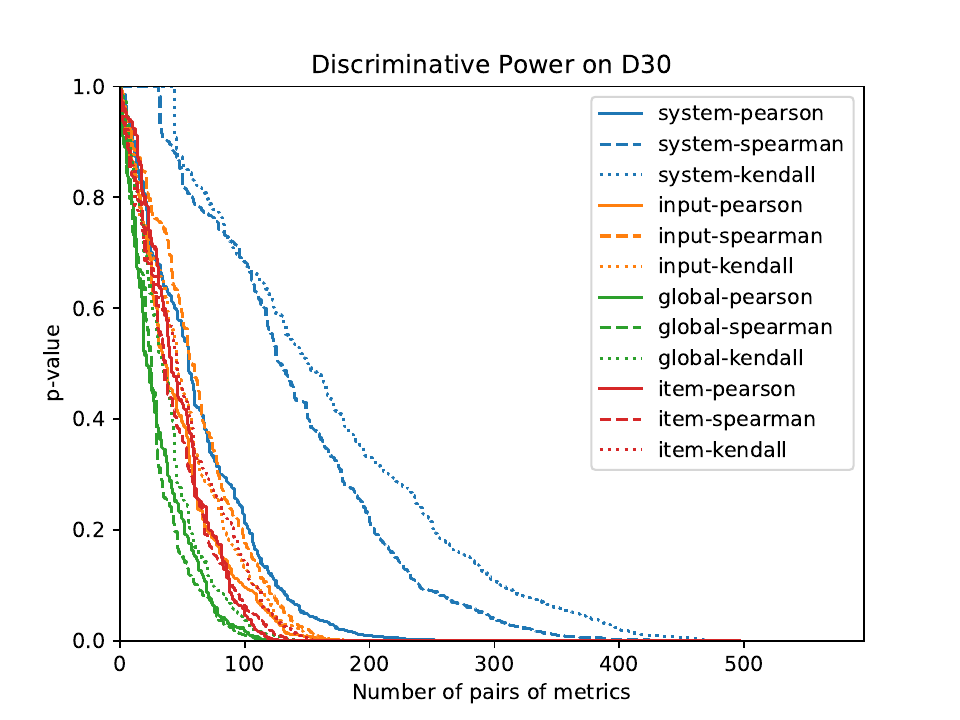} \hfill
  \caption {The p-value curves of correlation measures on meta-evaluation D30.}
  \label{fig:p_value_curve_D30}
\end{figure}

\begin{figure}[!htp]
  \includegraphics[width=\linewidth]{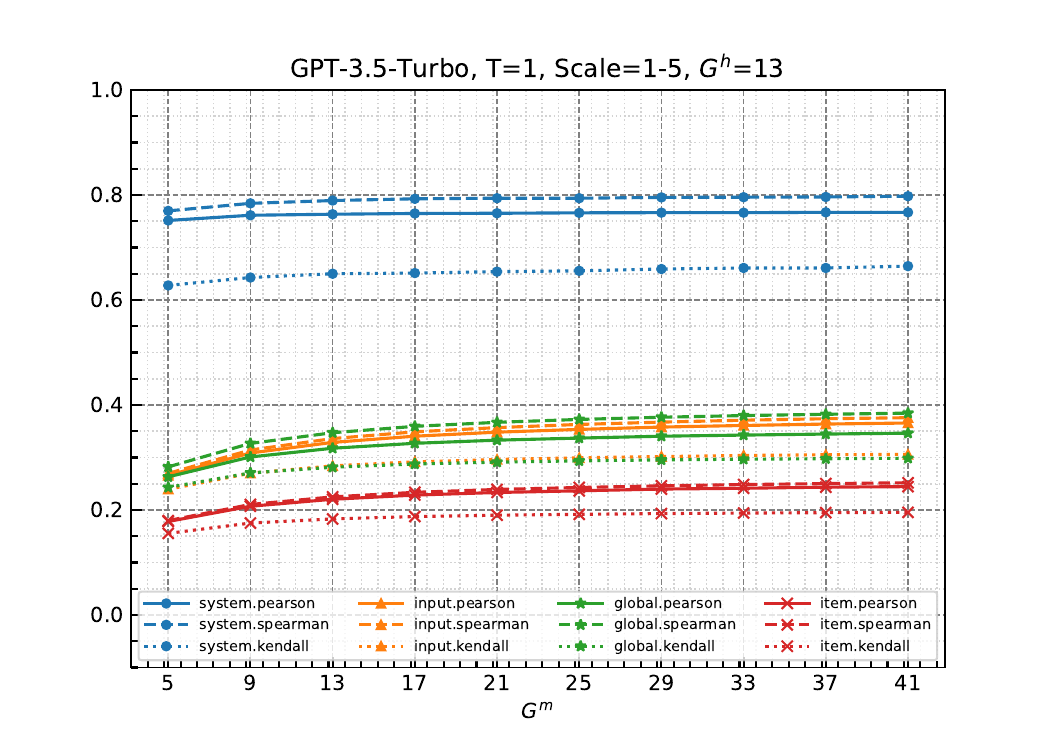}
  \caption{As the changes of $G^m$, the correlations between the GPT-3.5-Turbo evaluator and human evaluation using different measures on SummEval with the fixed evaluation scale of 1-5.}
  \label{fig:real1}
\end{figure}

\begin{figure}[!htp]
  \includegraphics[width=\linewidth]{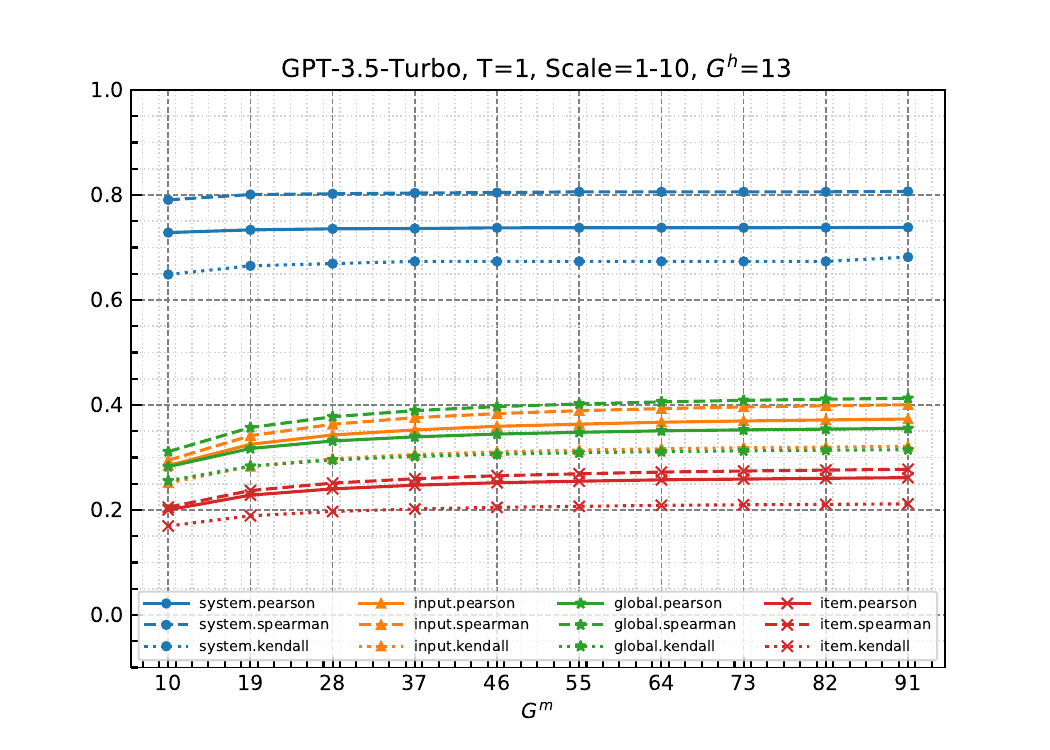}
  \caption{As the changes of $G^m$, the correlations between the GPT-3.5-Turbo evaluator and human evaluation using different measures on SummEval with the fixed evaluation scale of 1-10.}
  \label{fig:real2}
\end{figure}

\begin{figure}[!htp]
  \includegraphics[width=\linewidth]{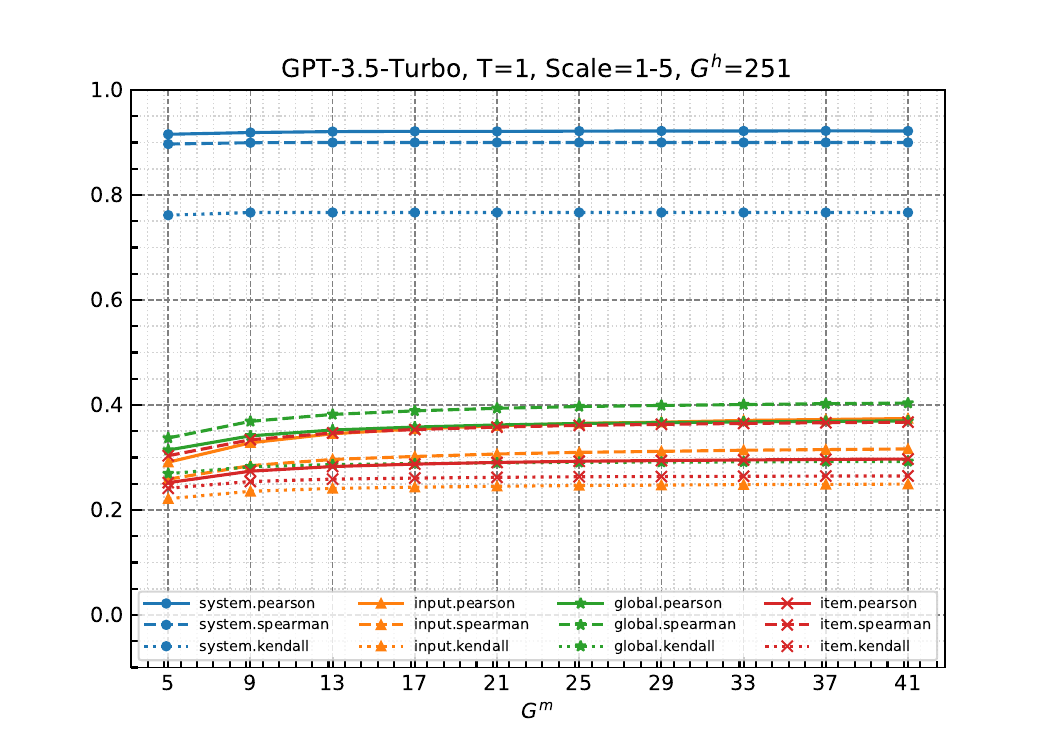}
  \caption{As the changes of $G^m$, the correlations between the GPT-3.5-Turbo evaluator and human evaluation using different measures on WMT23 with the fixed evaluation scale of 1-5.}
  \label{fig:real3}
\end{figure}

\begin{figure}[!htp]
  \includegraphics[width=\linewidth]{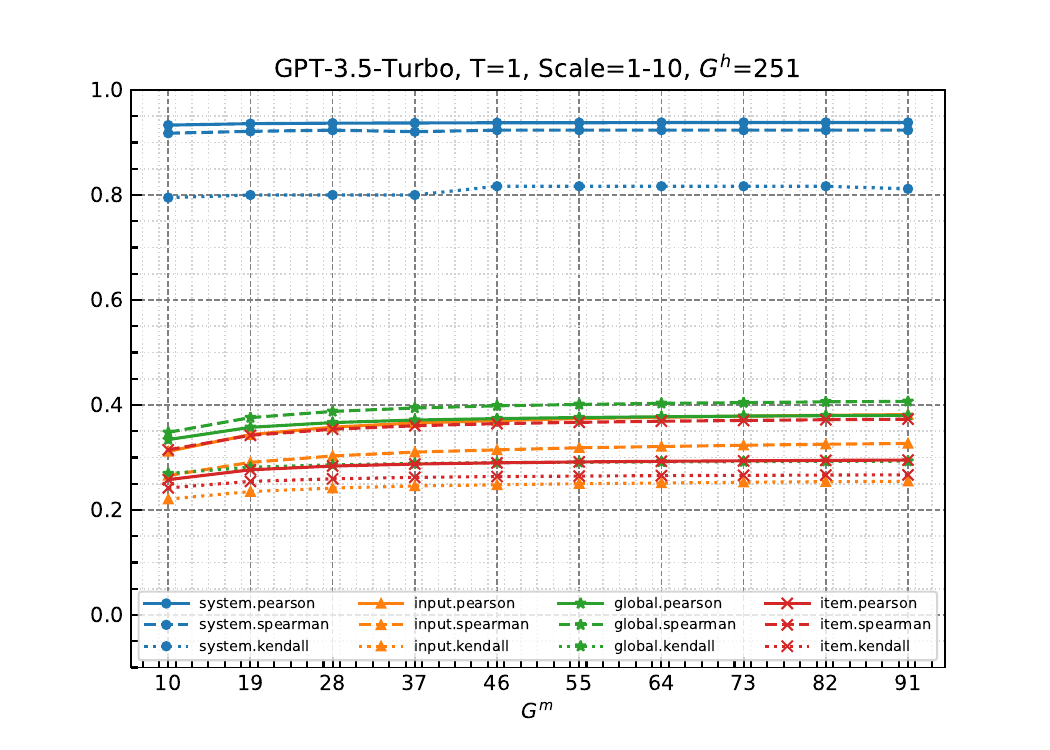}
  \caption{As the changes of $G^m$, the correlations between the GPT-3.5-Turbo evaluator and human evaluation using different measures on WMT23 with the fixed evaluation scale of 1-10.}
  \label{fig:real4}
\end{figure}

\begin{figure}[!htp]
  \includegraphics[width=\linewidth]{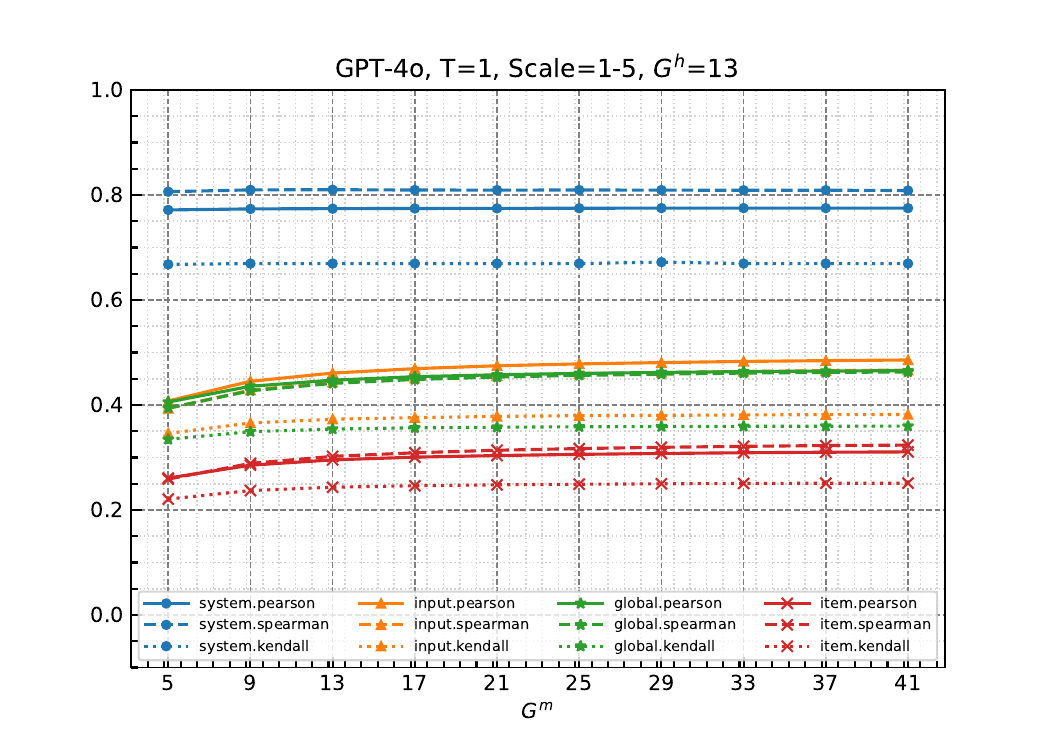}
  \caption{As the changes of $G^m$, the correlations between the GPT-4o evaluator and human evaluation using different measures on SummEval with the fixed evaluation scale of 1-5.}
  \label{fig:real5}
\end{figure}

\begin{figure}[!htp]
  \includegraphics[width=\linewidth]{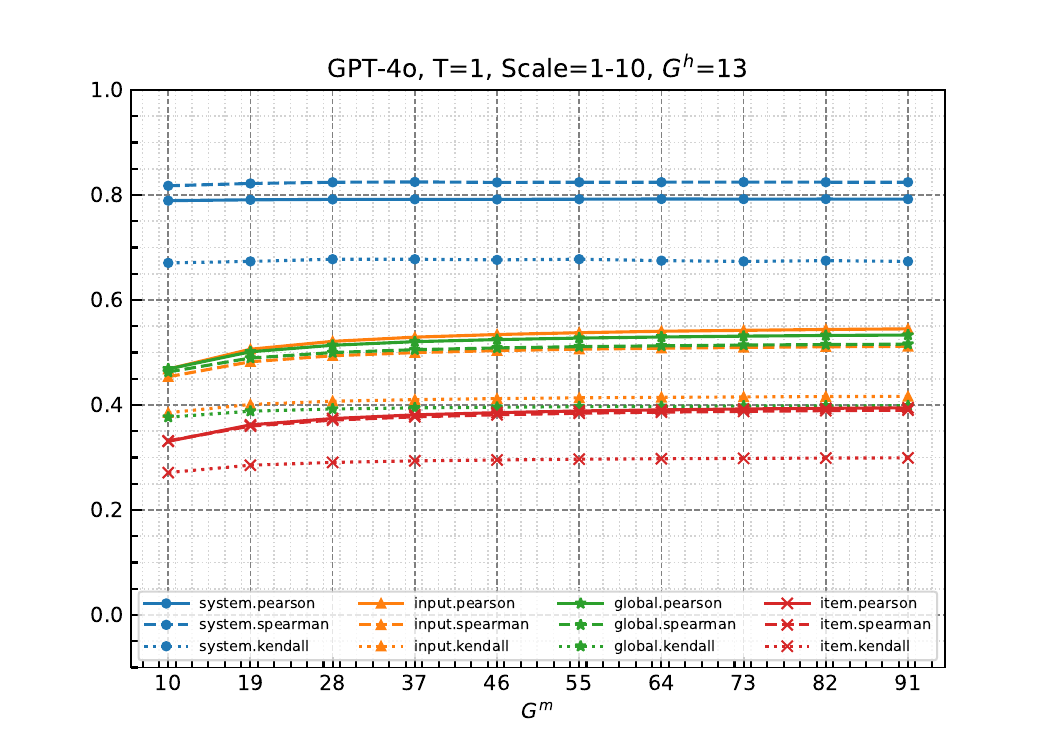}
  \caption{As the changes of $G^m$, the correlations between the GPT-4o evaluator and human evaluation using different measures on SummEval with the fixed evaluation scale of 1-10.}
  \label{fig:real6}
\end{figure}

\begin{figure}[!htp]
  \includegraphics[width=\linewidth]{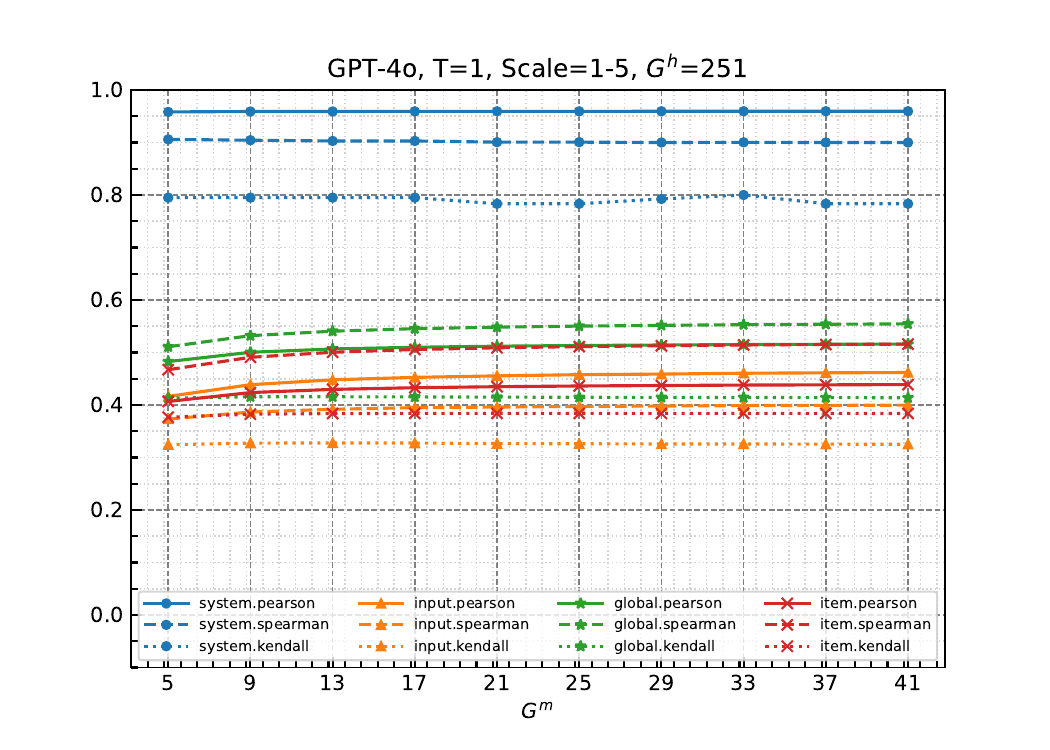}
  \caption{As the changes of $G^m$, the correlations between the GPT-4o evaluator and human evaluation using different measures on WMT23 with the fixed evaluation scale of 1-5.}
  \label{fig:real7}
\end{figure}

\begin{figure}[!htp]
  \includegraphics[width=\linewidth]{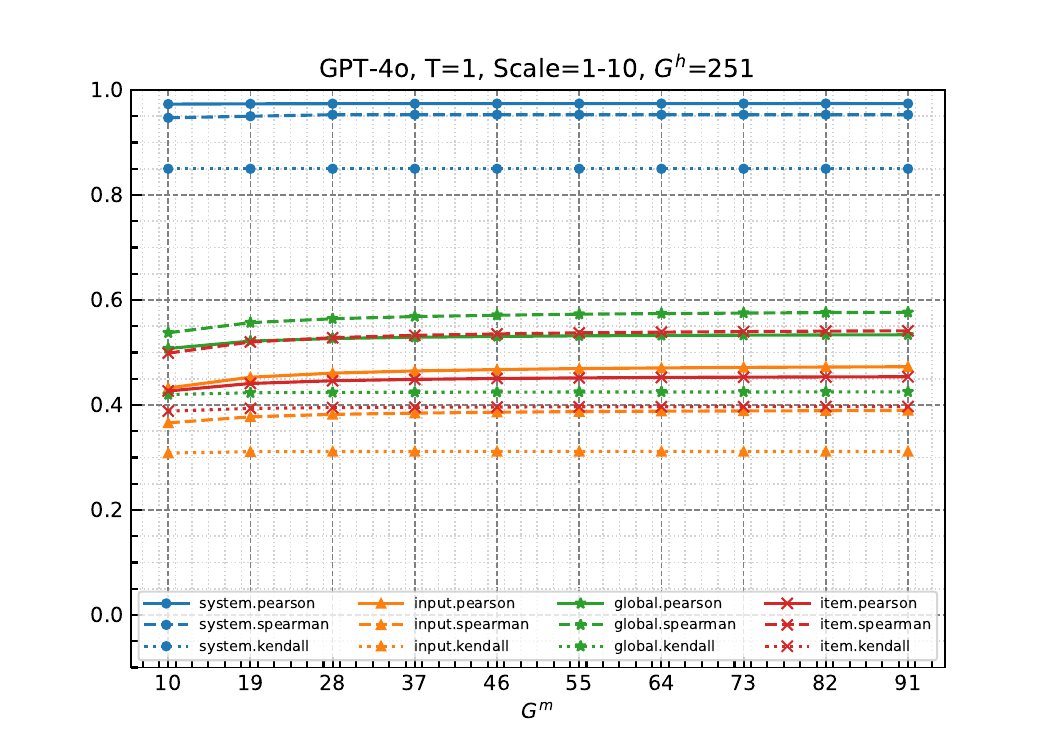}
  \caption{As the changes of $G^m$, the correlations between the GPT-4o evaluator and human evaluation using different measures on WMT23 with the fixed evaluation scale of 1-10.}
  \label{fig:real8}
\end{figure}

\begin{figure}[!htp]
  \includegraphics[width=\linewidth]{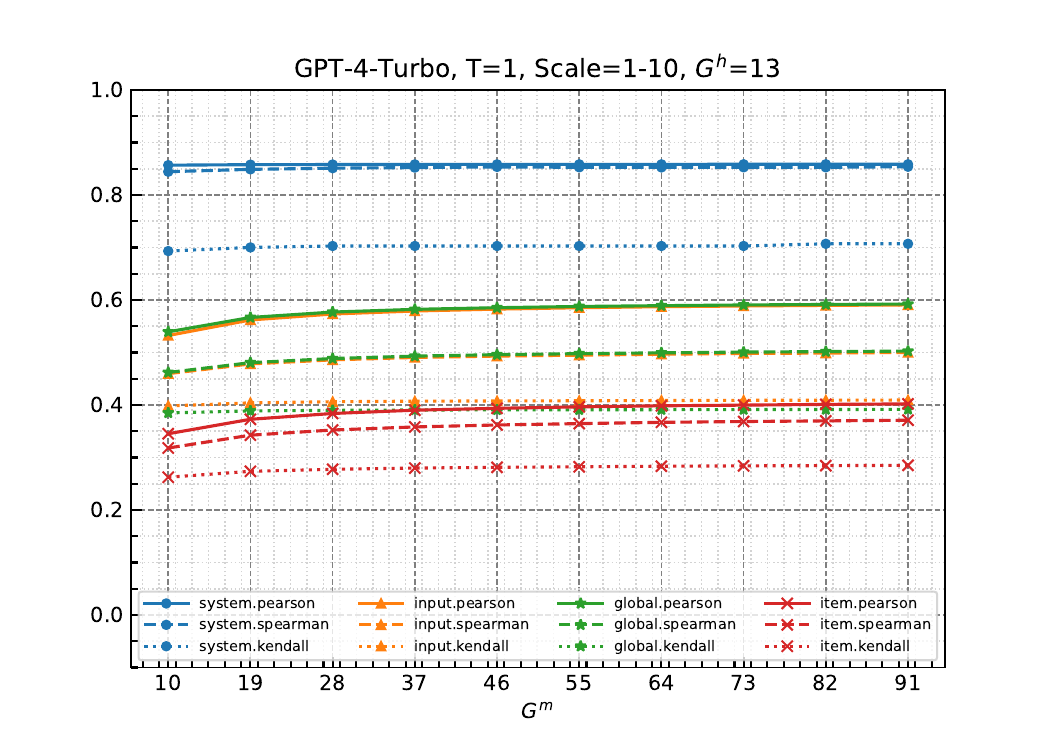}
  \caption{As the changes of $G^m$, the correlations between the GPT-4-Turbo evaluator and human evaluation using different measures on SummEval with the fixed evaluation scale of 1-10.}
  \label{fig:real9}
\end{figure}

\begin{figure}[!htp]
  \includegraphics[width=\linewidth]{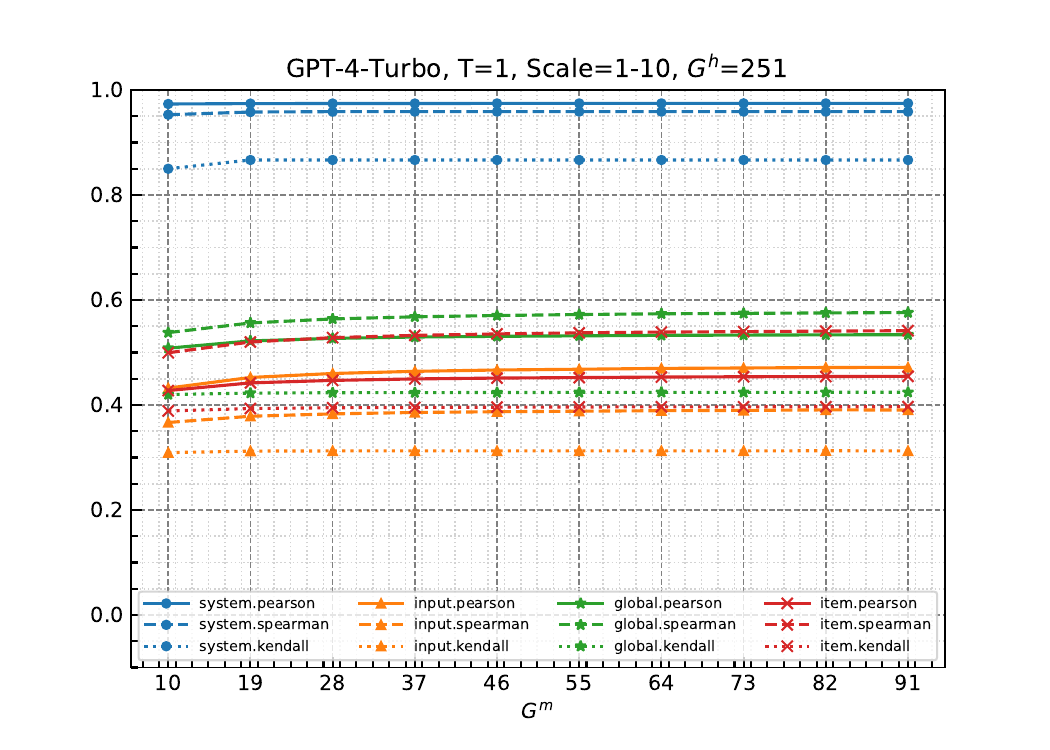}
  \caption{As the changes of $G^m$, the correlations between the GPT-4-Turbo evaluator and human evaluation using different measures on WMT23 with the fixed evaluation scale of 1-10.}
  \label{fig:real10}
\end{figure}

\begin{figure}[!htp]
  \includegraphics[width=\linewidth]{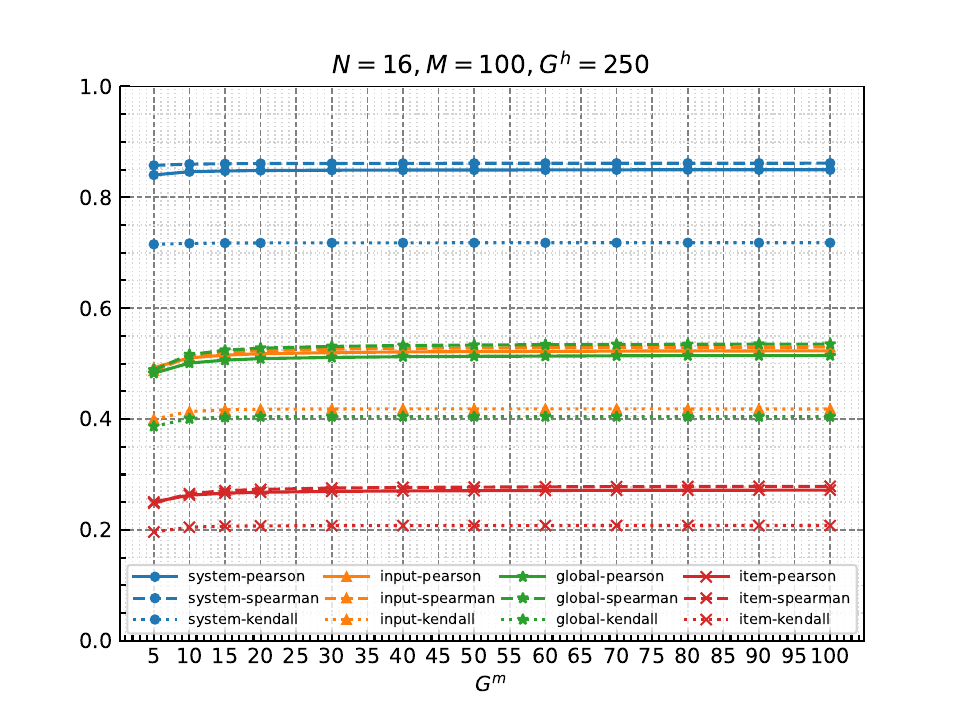}
  \caption{As the changes of $G^m$, the correlations between metrics and humans using different measures in statistical simulation with $G^h=250$.}
  \label{fig:simulation1}
\end{figure}

\end{document}